\documentclass[journal]{IEEEtran}
\ifCLASSINFOpdf
\else
\fi
\usepackage{hyperref}
\usepackage{cite}
\usepackage{graphicx} 
\usepackage{epstopdf}
\usepackage{amsfonts}
\usepackage{amsmath}
\usepackage{multicol}  
\usepackage{multirow}  
\usepackage{color}
\usepackage{subfigure}
\definecolor{ColorName}{rgb}{0,0,0}
\usepackage{xcolor}
\usepackage[normalem]{ulem} % use normalem to protect \emph
\newcommand\hl{\bgroup\markoverwith
	{\textcolor{yellow}{\rule[-.5ex]{2pt}{2.5ex}}}\ULon}
\usepackage[framemethod=tikz]{mdframed}
\usepackage{lipsum}
\begin{document}

\title{Domain-invariant Similarity Activation Map Contrastive Learning for  Retrieval-based Long-term  \\ Visual Localization}

\author{Hanjiang~Hu, Hesheng~Wang$^{*}$,~\IEEEmembership{Senior~Member,~IEEE,} Zhe~Liu, Weidong~Chen,~\IEEEmembership{Member,~IEEE}%
        %and~Jane~Doe,~\IEEEmembership{Life~Fellow,~IEEE}% <-this % stops a space
\thanks{This work was supported in part by the Natural Science Foundation of China under Grant U1613218, 61722309 and U1813206, in part by State Key Laboratory of Robotics and System (HIT). Corresponding Author: Hesheng Wang.   }% <-this % stops a space
\thanks{H. Hu, H. Wang and W. Chen are with Autonomous Robot Lab, Department of Automation, Key Laboratory of System Control and Information Processing of Ministry of Education, Shanghai Jiao Tong University, Shanghai 200240, China. Hesheng Wang is also with the State Key Laboratory of Robotics and System (HIT). Z. Liu is with the Department of Computer Science and Technology, University of Cambridge, United Kingdom. } }% <-this % stops a space
\maketitle

% The combination of To retrieve the target image in the database, a query image is encoded through encoder of its domain to obtain a domain-invariant feature, which is used to retrieve the most similar one in the database domain.
\begin{abstract}
	Visual localization is a crucial component in the application of mobile robot and autonomous driving. Image retrieval is an efficient and effective technique in image-based localization methods.  Due to the drastic variability of environmental conditions, \textit{e.g.} illumination, seasonal and weather changes, retrieval-based visual localization is severely affected and becomes a challenging problem. In this work, a general architecture is first formulated probabilistically to extract domain-invariant feature through multi-domain image translation. And then a novel gradient-weighted similarity activation mapping loss (Grad-SAM) is incorporated for finer localization with high accuracy. We also propose a new adaptive triplet loss to boost the contrastive learning of the embedding in a self-supervised manner. The final coarse-to-fine image retrieval pipeline is implemented as the sequential combination of models without and with Grad-SAM loss.  Extensive experiments have been conducted to validate the effectiveness of the proposed approach on the CMU-Seasons dataset. The strong generalization ability of our approach is verified on RobotCar dataset using models pre-trained on urban part of CMU-Seasons dataset. Our performance is on par with or even outperforms the state-of-the-art image-based localization baselines in medium or high precision, especially under the challenging environments with illumination variance, vegetation and night-time images. The code and  pretrained models are available on \url{https://github.com/HanjiangHu/DISAM}.
\end{abstract}


% \fi
\IEEEpeerreviewmaketitle

\section{Introduction}
\IEEEPARstart{V}{isual} localization is an essential problem in visual perception of autonomous driving or mobile robots \cite{sattler2016efficient,bescos2018dynaslam, wang2020avoiding}, which is low-cost and efficient compared with Global Positioning System-based (GPS-based) or Light Detection and Ranging-based (LiDAR-based) localization methods. Image retrieval,  \textit{i.e.} recognizing the most similar place in the database for each query image \cite{arandjelovic2017netvlad,lowry2015visual,torii201824}, is a convenient and effective technique for image-based localization, \textcolor{ColorName}{ which both serves as place recognition for loop closure in Simultaneous Localization and Mapping (SLAM) and provides initial pose for finer 6-DoF camera pose regression \cite{sattler2019understanding, sarlin2019coarse}.}

\begin{figure}[htbp]
	
	\centering
	
	\subfigure[]{
		\begin{minipage}[t]{0.5\linewidth}
			\centering
			\includegraphics[width=1.52in]{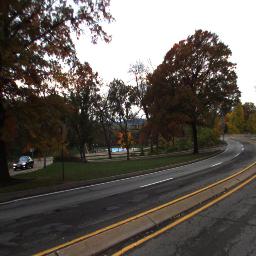}\\
			\vspace{0.1cm}
			\includegraphics[width=1.52in]{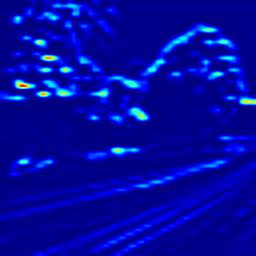}\\
			\vspace{0.1cm}
			%\caption{fig1}
		\end{minipage}%
	}%
	\subfigure[]{
		\begin{minipage}[t]{0.5\linewidth}
			\centering
			\includegraphics[width=1.52in]{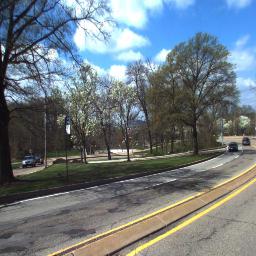}\\
			\vspace{0.1cm}
			\includegraphics[width=1.52in]{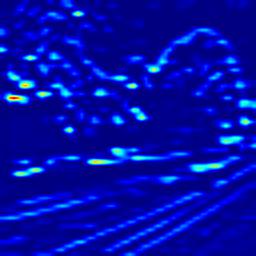}\\
			\vspace{0.1cm}
			%\caption{fig1}
		\end{minipage}%
	}%

	\centering
%	\vspace{-0.2cm}
	\caption{On the first row, Column (a) shows a query image under \textit{Overcast + Mixed Foliage} condition and Column (b) shows the retrieved image under \textit{Sunny + No Foliage} condition. On the second row, the gradient-weighted similarity activation maps are shown for the above images. The activation map visualizes the salient area on the image which contributes most to the matching and retrieval across the different environments. }
%	\vspace{-0.5cm}
	\label{demo_fig}
\end{figure}

However, the drastic perceptual changes caused by long-term environmental condition variance, \textit{e.g.} changing seasons, illumination and weather, cast serious challenges on image-based localization in long-term outdoor self-driving scenarios \cite{sattler2018benchmarking}. Traditional feature descriptors (SIFT, BRIEF, ORB, BRISK, \textit{etc.}) could be only used for image matching under scenes without significant appearance changes due to the reliance on image pixels. With convolutional neural networks (CNNs) making remarkable progress in the field of computer vision and autonomous driving \cite{ma2020artificial}, learning-based methods have been paid significant attention to owing to the robustness of deep features against changing environments for place recognition and retrieval \cite{doan2019scalable,yin2019multi,chen2017deep}.

\textcolor{ColorName}{Contrastive learning is an important technique for image recognition tasks \cite{wang2014learning,wohlhart2015learning,lu2017discriminative}, also known as deep metric learning, which aims to learn metrics and latent representations with closer distance for similar images. Compared to face recognition, supervised learning for place recognition \cite{lowry2016supervised,chen2017deep} suffers from difficulty in determining which clip of images should be grouped to the same place in the sequence of continuous images. Moreover, supervised contrastive learning methods for outdoor place recognition \cite{gordo2017end,radenovic2018fine} need numerous paired samples for model training due to heterogeneously entangled scenes with multiple environmental conditions, which is effort-cost and inefficient. Additionally, considering the feature map with salient areas in the explanation of CNNs for classification task \cite{zhou2016learning,selvaraju2017grad,chattopadhay2018grad}, retrieval-based localization could be addressed through such attentive or contextual information \cite{kim2017learned,chen2018learning}. However, these methods have no direct access to the similarity of the extracted feature so they are not appropriate for high-precision localization. }

To address these issues, we first propose  an unsupervised and implicitly content-disentangled representation learning through  probabilistic modeling to obtain domain-invariant features (DIF) based on multi-domain image translation with feature consistency loss (FCL). For the retrieval with high accuracy, a novel gradient-weighted similarity activity mapping (Grad-SAM) loss is introduced inside the training framework inspired by \cite{zhou2016learning,selvaraju2017grad,chattopadhay2018grad}. Furthermore, a novel unsupervised adaptive triplet loss is incorporated in the pipeline to promote the training of FCL or Grad-SAM \textcolor{ColorName}{and the two-stage test pipeline is implemented in a coarse-to-fine manner for the performance compensation and improvement.} We further investigate the localization and place recognition performance of the proposed method by conducting extensive experiments on both CMU-Seasons dataset and RobotCar-Seasons dataset. Compared to state-of-the-art image-based baselines, our method presents competitive results in medium and high precision. The example of image retrieval is shown in Figure \ref{demo_fig} . Our contributions are summarized as follows:

\begin{itemize}
	\item A domain-invariant feature learning framework is proposed based on multi-domain image-to-image translation architecture with feature consistency loss and is statistically formulated as a probabilistic model of image disentanglement.
	\item A new gradient-weighted similarity activation mapping (Grad-SAM) loss is proposed inside the framework to leverage the localizing information of feature map for high-accuracy retrieval.
	\item A novel adaptive triplet loss is introduced for FCL or Grad-SAM learning for the self-supervised contrastive learning and give the effective two-stage retrieval pipeline from coarse to fine.
	\item The effectiveness and strong generalization ability of the proposed approach is validated on CMU-Seasons dataset and RobotCar-Seasons dataset for visual localization through extensive experiments. Ours results keep on par with state-of-the-art  baselines of image retrieval-based localization for medium and high precision.
\end{itemize}

The rest of this paper is organized as follows. Section \hyperref[sec2]{\ref{sec2}} presents the related work in place recognition and representation learning for image retrieval. Section \hyperref[sec3]{\ref{sec3}} presents the formulation of domain-invariant feature learning model with FCL. Section \hyperref[sec4]{\ref{sec4}} introduces the adaptive triplet loss and the two-stage retrieval pipeline with Grad-SAM loss. Section \hyperref[sec5]{\ref{sec5}} shows the experimental results on visual localization benchmark. Finally, in Section \hyperref[sec6]{\ref{sec6}} we draw our conclusions and present some suggestions for the  future work.

\section{Related Work}
\label{sec2}

\subsection{Place Recognition and Localization}
\label{sec2sub1}

Outdoor visual place recognition has been studied for many years and could be directly used for visual localization in autonomous driving or loop closure detection of SLAM, in which the most similar images are retrieved from database for query images. \textcolor{ColorName}{Traditional local feature descriptors are aggregated for image retrieval \cite{jegou2010aggregating,galvez2012bags,shi2019mixed,cummins2008fab}, and have successfully addressed most cases of loop closure detection in real-time visual SLAM \cite{mur2017orb,milford2012seqslam} without significant environmental changes.} VLAD \cite{jegou2010aggregating} is the most successful man-made feature for place recognition and has been extended to different versions. NetVLAD\cite{arandjelovic2017netvlad} extract deep features through VLAD-like network architecture.  DenseVLAD \cite{torii201824} presents impressive results through extracting multi-scale SIFT descriptor for VLAD\cite{jegou2010aggregating} under drastic perceptual variance.

Since convolutional neural networks (CNNs) has successfully addressed many tasks in computer vision \cite{xing2018advances}, long-term visual place recognition and localization have significantly developed  assisted with CNNs \cite{arandjelovic2017netvlad,chen2017deep,jenicek2019no}. To cope with the challenging perceptual changes, many recent works follow the pipeline of learning the robust deep representation through neural networks together with semantic \cite{naseer2017semantics,stenborg2018long}, geometric\cite{piasco2019learning}, context-aware information\cite{kim2017learned,chen2018learning,xin2019localizing}, \textit{etc.}. \textcolor{ColorName}{Although attentive information could be  learned through biological  visual  attention \cite{wang2017hierarchical},} most of these methods mainly adopt end-to-end training manner, the representation features are directly trained as global place-specific features with the aid of auxiliary information which is effort-cost to obtain. Inspired by classification activation map \cite{zhou2016learning,selvaraju2017grad,chattopadhay2018grad} in visual explanation of CNN, we introduce the notion of activation map to image retrieval problem, which could be intuitively regarded as the classification problem of infinite place classes by replacing the class score with the similarity of image pairs.

Another solutions to the change of appearance are based on image translation\cite{isola2017image,anoosheh2018combogan,liu2017unsupervised,huang2018multimodal}, where images are transfered across different domains based on generative adversarial networks (GANs) \cite{goodfellow2014generative,radford2015unsupervised}. Porav \textit{et al.} \cite{porav2018adversarial} first translates query images to database domain through CycleGAN\cite{zhu2017unpaired} and retrieves target images through hand-crafted descriptors. ToDayGAN \cite{anoosheh2019night} similarly translates night-images to day-images and uses DenseVLAD for retrieval. Jenicek \textit{et al.} \cite{jenicek2019no} proposes to use U-Net to obtain photometric normalization image and finds deep embedding for retrieval. However, generalization ability is limited by translation-based methods because the accuracy of retrieval on image level  largely depends on the quality of translated image compared to the retrieval with latent-feature .

\subsection{Disentanglement Representation}
\label{sec2sub2}
\textcolor{ColorName}{Latent representation reveals the feature vectors in the latent space which determine the distribution of samples. Therefore, it is essential to find the latent disentangled representation to analyze the attributes of data distribution. A similar application is the Latent Factor Model (LFM) in recommender systems \cite{luo2019latent,8941240,8802269}, where the latent factor contributes to the preference of specific users.} In the field of style transfer or image translation \cite{isola2017image,gong2019dlow}, the deep representations of images are modeled according to the variations of data which depend on different factors across domains  \cite{achille2018emergence,kim2018disentangling}, \textit{e.g.} disentangled content and style representation. Supervised approaches \cite{makhzani2015adversarial,mathieu2016disentangling} learn class-specific representations through labeled data, and many works have appeared to learn disentangled representation in unsupervised manner\cite{chen2016infogan,donahue2018semantically}. Recently, fully- and partially-shared representation of latent space have been investigated for unsupervised image-to-image translation\cite{liu2017unsupervised,huang2018multimodal,lee2018diverse}. Inspired by these methods, where the content code is shared across all the domains but the style code is domain-specific, our domain-invariant representation learning is probabilistically formulated and modeled as an extended and modified version of CycleGAN\cite{zhu2017unpaired} or ComboGAN \cite{anoosheh2018combogan}.

For the application of representation learning in place recognition under changing environments, where each environmental condition corresponds to one domain style and the images share similar scene content across different environments, it is appropriate to make the assumption of disentangled representation to this problem case. Recent works for condition-invariant deep representation learning \cite{lowry2015visual,lopez2017appearance,hu2019retrieval} in long-term changing environments mainly rely on variance-removal or other auxiliary information introduced in \hyperref[sec2sub1]{\ref{sec2sub1}} . \cite{lowry2016supervised} removes the dimension related to the changing condition through PCA for the deep embeddings of latent space through classification model. \cite{yin2019multi} separates the condition-invariant representation from VLAD features with GANs across multiple domains. \cite{hausler2019filter} filters the distracting feature maps in the shallow CNNs but matches with deep features in deeper CNNs to improve condition- and viewpoint-invariance \cite{garg2018don} using image pairs. Compared to these two-stage or supervised methods, an end-to-end domain-invariant feature learning method \cite{hu2019retrieval} possesses advantages on direct, low-cost and efficient learning.

\subsection{Contrastive Learning}
\label{sec2sub3}
% first background, them come to topic
Contrastive learning, a.k.a., deep metric learning \cite{wang2014learning,oh2016deep} stems from distance metric learning \cite{xing2003distance,weinberger2008fast} in machine learning but extracts deep features through deep neural networks, \textit{i.e.} learning appropriate embeddings and metrics for effective discrimination between similar sample pairs and different sample pairs. With the help of neural networks, deep metric learning typically utilizes siamese network \cite{varior2016gated,balntas2018relocnet} or triplet network\cite{hoffer2015deep,kumar2016learning}, which makes the embedding of same category closer than that of different category with triple labeled input samples for face recognition, human re-identification, \textit{etc.}.

Coming to long-term place recognition and visual localization, recently many works use supervised learning together with siamese network and triplet loss \cite{gordo2017end,lopez2017appearance}. To avoid vanishing gradient of small distance from different pairs with triplet loss form \cite{wang2014learning},  \cite{wohlhart2015learning} proposes another form of triplet loss.  Due to the hard-annotated data for supervised learning, Radenovic \textit{et al.} \cite{radenovic2018fine} proposes to leverage geometry of 3D model from structure-from-motion (SfM) for triplet learning in an automated manner. But SfM is off-line and effort-cost so it is not possible for end-to-end training. Instead we employ unsupervised triplet training technique adapted to the DIFL framework \cite{hu2019retrieval} so that domain-invariant and scene-specific representation could be trained in unsupervised and end-to-end way efficiently.

\section{Formulation of Domain-Invariant Feature Learning}
\label{sec3}

\subsection{Problem Assumptions}
\label{sec3sub1}

Our approach to long-term visual place localization and recognition is modeled in the setting of multi-domain unsupervised image-to-image translation, where all query and database images are captured from multiple identical sequences across environments. Images in different environmental conditions belong to corresponding domains respectively. Let the total number of domains be denoted as $ N $ and two different domains are randomly sampled from $ \{1 \cdots N\} $ for each translation iteration, \textit{e.g.} $ i,j \in \{1 \cdots N\}, i\neq j $. Let $ x_{i} \in \mathcal{X}_{i} $ and $ x_{j} \in \mathcal{X}_{j} $ represent images from these two domains. For the multi-domain image-to-image translation task \cite{anoosheh2018combogan}, the goal is to find all conditional distributions $ p(x_{i}|x_{j}), \forall i\neq j, i,j \in \{1 \cdots N\} $ with known marginal distribution of $ p(x_{i}), p(x_{j})$ and translated conditional distribution $ p(x_{j \to i}|x_{j}), p(x_{i \to j}|x_{i})$. Since different domains correspond to different environmental conditions, we suppose the conditional distribution $ p(x_{i}|x_{j}) $ is monomodal and deterministic compared to multimodal distribution across only two domains in \cite{huang2018multimodal}. As $ N $ increases to infinity and becomes continuous, the multi-domain translation model covers more domains and could be regarded as a generalized multi-modal translation with limited domains.

Like the shared-latent-space assumption in the recent unsupervised image-to-image translation methods \cite{liu2017unsupervised,huang2018multimodal,lee2018diverse}, the content representation $ c $ is shared  across different domains while the style latent variable $ s_{i} $ belongs to each specific domain. For the image joint distribution in one domain $ x_{i} \in \mathcal{X}_{i} $, it is generated from the prior distribution of content and style,  $ x_{i} = G_{i}(s_{i},c) $, and the content and style are independent to each other. Since the condition distribution $ p(x_{i}|x_{j}) $ is deterministic, the style variable is only embodied in the latent generator of the specific domain, \textit{i.e.} $ x_{i} = G_{i}(s_{i},c) = D_{i}^{c*}(c), x_{j} = G_{j}(s_{j},c) = D_{j}^{c*}(c) $. Under such assumptions, our method could be regarded as implicitly partially-shared although only content latent code is explicitly found across multiple domains with corresponding generators. Following the previous work \cite{huang2018multimodal}, we further assume that the domain-specific decoder functions for shared content code, $ D_{i}^{c*},D_{j}^{c*} $, are deterministic and their inverse encoder functions exist, where $ E_{i}^{c*}=(D_{i}^{c*})^{-1} ,E_{j}^{c*}=(D_{j}^{c*})^{-1} $. And our goal of domain-invariant representation learning is to find the underlying decoders $ D_{i}^{c*},D_{j}^{c*} $ and encoders $ E_{i}^{c*},E_{j}^{c*} $ for all the environmental domains through neural networks, so that the domain-invariant latent code $ c $ could be extracted for any given image sample $ x_{i} $ through $ c=E_{i}^{c*}(x_{i}) $.
% different from MUNIT because the style is not explicit variable but embodied in the encoder and docoder of each domain

\begin{figure}[thpb]
	\centering
	%\framebox
	%{
	\includegraphics[scale=0.43]{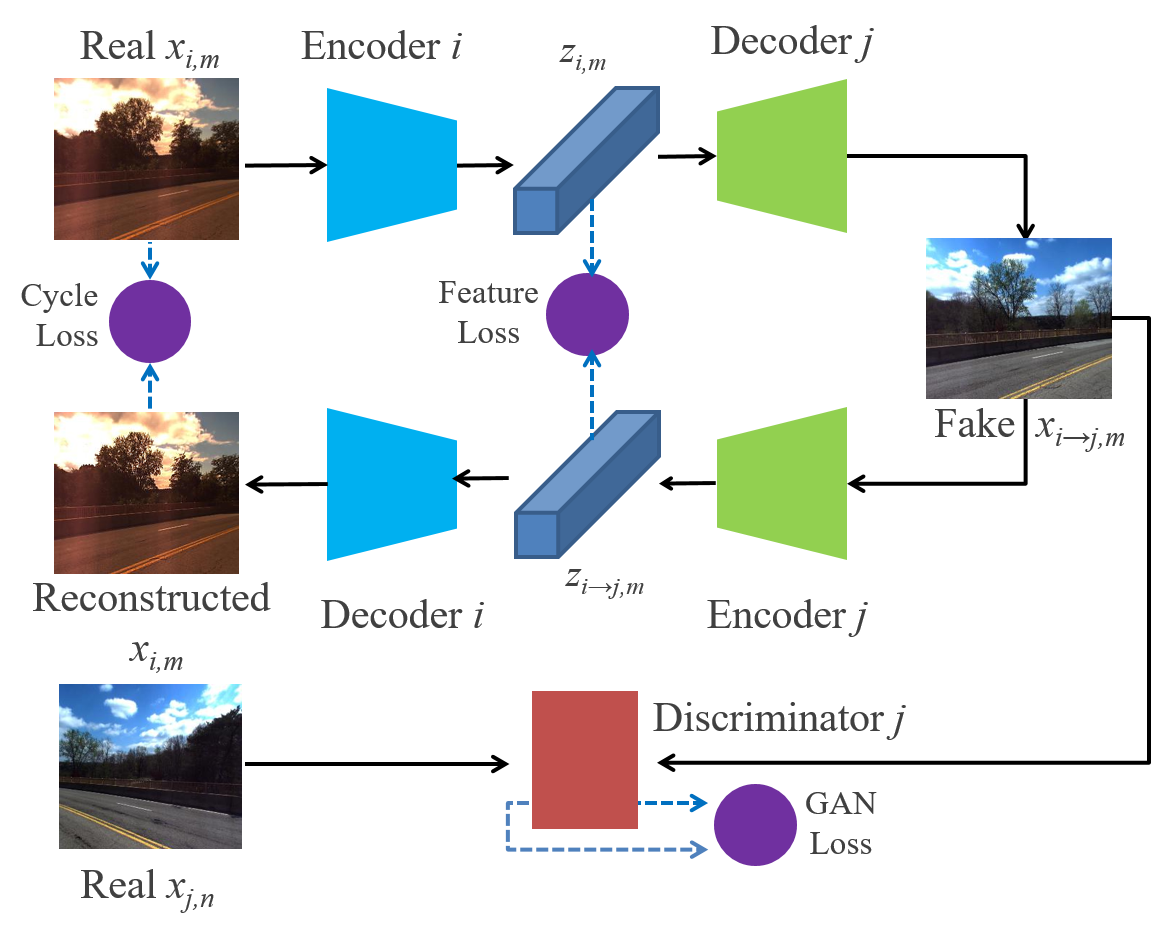}
	%\begin{mdframed}[hidealllines=true,backgroundcolor=blue!20]
		%\lipsum[2]
	%	\includegraphics[scale=0.43]{DIFL.png}
	%\end{mdframed}
	
%	\vspace{-0.5cm}
	%}
	
	\caption
	{
		Network architecture for image translation from domain $ i $ to $ j $. Constraint by GAN loss, cycle loss and feature loss, the latent feature code is the domain-invariant representation. \textcolor{ColorName}{The discriminator $ D_j $ results in GAN loss through adversarial training, given the real image of domain $ j $ and the translated image from domain $ i $ to $ j $.}
	}
%	\vspace{-0.5cm}
	\label{DIFL_fig}
\end{figure}

\subsection{Model Architecture}
\label{sec3sub2}
% introduce the architecture and build the probablistic model
% list the loss and show the case cannot be DIF

We adopt the multi-domain image-to-image translation architecture \cite{anoosheh2018combogan}, which is an expansion of CycleGAN \cite{zhu2017unpaired} from two domains to multiple domains. The generator networks in the framework are decoupled into domain-specific pairs of encoders $ E_{i}^{c} $ and decoders $ D_{i}^{c} $ for any domain $ i $. The encoder are the first half of the generator while the decoder is the second half for each domain. For the image translation across multiple domains, the encoders and decoders could be randomly combined like manipulation of blocks. The discriminators $ D_{i} $ are also domain-specific for domain $ i $ and optimized in adversarial training as well. \textcolor{ColorName}{The detailed architectures of encoder, decoder and discriminator for each domain is the same as ComboGAN \cite{anoosheh2018combogan}.}

For images in similar sequences under different environments, first suppose domain $ i,j $ are selected randomly and images are denoted as $ x_{i},x_{j} $.  The basic framework DIFL is shown as Figure \ref{DIFL_fig} , including GAN loss, cycle consistency loss and feature consistency loss. For the image translation pass from domain $ i $ to domain $ j $, the latent feature is first encoded by encoder $ E_{i}^{c} $ and then decoded by decoder $ D_{j}^{c} $. The translated image goes back through encoder $ E_{j}^{c} $ and decoder $ D_{i}^{c} $ to find the cycle consistency loss (\hyperref[cycle_loss]{\ref{cycle_loss}}) \cite{zhu2017unpaired}. Also, the translated image goes through the discriminator $ D_{j} $ to find adversarial loss (\hyperref[gan_loss]{\ref{gan_loss}}) \cite{goodfellow2014generative}. The pass from domain $ j $ to domain $ i $ is similar.
\begin{eqnarray}
\label{cycle_loss}    
\mathcal{L}_{Cycle}^{x_{i}}=\mathbb{E}_{x_{i} \sim p(x_{i})}[\|(D_{i}^{c}(E_{j}^{c}((D_{j}^{c}(E_{i}^{c}(x_{i}))))))-x_{i}\|_{1}]
\end{eqnarray}
The adversarial loss (\hyperref[gan_loss]{\ref{gan_loss}}) makes the translated image $ x_{i\to j} $ indistinguishable from the real image $ x_{j}$  and the distribution of translated images close to the ditribution of real images.
\begin{eqnarray}
\label{gan_loss}
\mathcal{L}_{GAN}^{x_{i}}=\mathbb{E}_{x_{j} \sim p(x_{j})}[(D_{j}(x_{j})-1)^{2}]\notag \\  +\mathbb{E}_{x_{i} \sim p(x_{i})}[D_{j}(D_{j}^{c}(E_{i}^{c}(x_{i})))^{2}]
\end{eqnarray}
The cycle consistency loss (\hyperref[cycle_loss]{\ref{cycle_loss}}) originates from CycleGAN \cite{zhu2017unpaired}, which has been proved to infer deterministic translation \cite{huang2018multimodal} and is suitable for representation learning through image translation among multiple domains.
For the pure multi-domain image translation task, \textit{i.e.} ComboGAN \cite{anoosheh2018combogan}, the total ComboGAN loss only contains adversarial loss and cycle consistency loss.
\begin{eqnarray}
\label{combogan_loss}    
\mathcal{L}_{ComboGAN}=\mathcal{L}_{GAN}^{x_{i}} + \mathcal{L}_{GAN}^{x_{j}} + \lambda_{cyc}(\mathcal{L}_{Cycle}^{x_{i}}+\mathcal{L}_{Cycle}^{x_{j}})
\end{eqnarray}

Since every domain owns a set of encoder, decoder and discriminator, the total architecture is complicated and could be modeled through probabilistic graph if all the encoders and decoders are regarded as conditional probability distribution. Supposing the optimality of ComboGAN loss (\hyperref[combogan_loss]{\ref{combogan_loss}}) is reached, the complex forward propagation during training could be simplified and the representation embedding could be analyzed.

Without loss of generality, image $ x_{i,m}, x_{j,n} $ are selected from image sequences $ x_{i}, x_{j}, i\neq j $,  where $ m,n $ represent the places of the shared image sequences and only related to the content of images. According to the assumptions in \hyperref[sec3sub1]{\ref{sec3sub1}} ,  $ m,n $ represent the shared domain-invariant content latent code $ c $ across different domains. For the translation from image $ x_{i,m} $ to domain $ j $, we have,
\begin{eqnarray}
\label{equ4}
&z_{i,m} = E_{i}^{c}(x_{i,m}) \\
\label{equ5}
&x_{i\to j,m} = D_{j}^{c}(z_{i,m})  \\
\label{equ6}
&z_{i\to j,m} = E_{j}^{c}(x_{i\to j,m}) \\
\label{equ7}
&x_{i,m} = D_{i}^{c}(z_{i \to j,m})
\end{eqnarray}
The latent code $ z_{i,m} $ implies the relationship of domain $ i $ and the content of image $ m $ from (\hyperref[equ4]{\ref{equ4}}). Due to the adversarial loss (\hyperref[gan_loss]{\ref{gan_loss}}), the translated image $ x_{i \to j,m} $ has the same distribution as image $ x_{j,n} $, \textit{i.e.} $ x_{i \to j, m}, x_{j,n} \sim p(x_{j}) $. For the reconstructed image from (\hyperref[equ7]{\ref{equ7}}), the cycle consistency loss (\hyperref[cycle_loss]{\ref{cycle_loss}}) limits it to the original image $ x_{i,m} $.

From Equation (\hyperref[equ4]{\ref{equ4}}), (\hyperref[equ5]{\ref{equ5}}), we have,
\begin{eqnarray}
\label{equ8}
	p(x_{i\to j,m})= p(x| z_{i,m}, D_{j}^{c}) = p(x_{j,m})
\end{eqnarray}
which indicates $ x_{i\to j,m} $ and $ i $ are independent if the optimality of adversarial loss (\hyperref[gan_loss]{\ref{gan_loss}}) is reached, and $ z_{i\to j,m} $ and $ i $ are also independent further from Equation  (\hyperref[equ6]{\ref{equ6}}). Similarly,  $ z_{i,m} $ and $ j $ are  independent for any $ j \neq i $.
Combine Equation (\hyperref[equ5]{\ref{equ5}}), (\hyperref[equ6]{\ref{equ6}}) and Equation (\hyperref[equ4]{\ref{equ4}}), (\hyperref[equ7]{\ref{equ7}}), we could find the relationship between $ z_{i,m} $ and $ z_{i\to j,m} $ and the weak form of inverse constraint on encoders and decoders below.
\begin{eqnarray}
\label{equ9}
&z_{i\to j,m} = E_{j}^{c}(D_{j}^{c}(z_{i,m})) \\
\label{equ10}
&z_{i,m} = E_{i}^{c}(D_{i}^{c}(z_{i\to j,m})) \\
\label{equ11}
&E_{i}^{c}(D_{i}^{c}) = (E_{j}^{c}(D_{j}^{c}))^{-1}, \forall i \neq j
\end{eqnarray}

When the optimality of original ComboGAN loss  (\hyperref[combogan_loss]{\ref{combogan_loss}}) is reached, for any $ i \neq j $, the latent code $ z_{i,m} $ and $ z_{i\to j,m} $ are not related to $ j $ and $ i $ respectively, which is consistent with the proposition that cycle consistency loss cannot infer shared-latent learning in \cite{liu2017unsupervised}. Consequently, the representation embeddings are not domain-invariant and not appropriate for image retrieval. And the underlying inverse encoders and decoders have not been found through vanilla ComboGAN image translation model.

\begin{figure}[thpb]
	\centering
	%\framebox
	%{
	\includegraphics[scale=0.4]{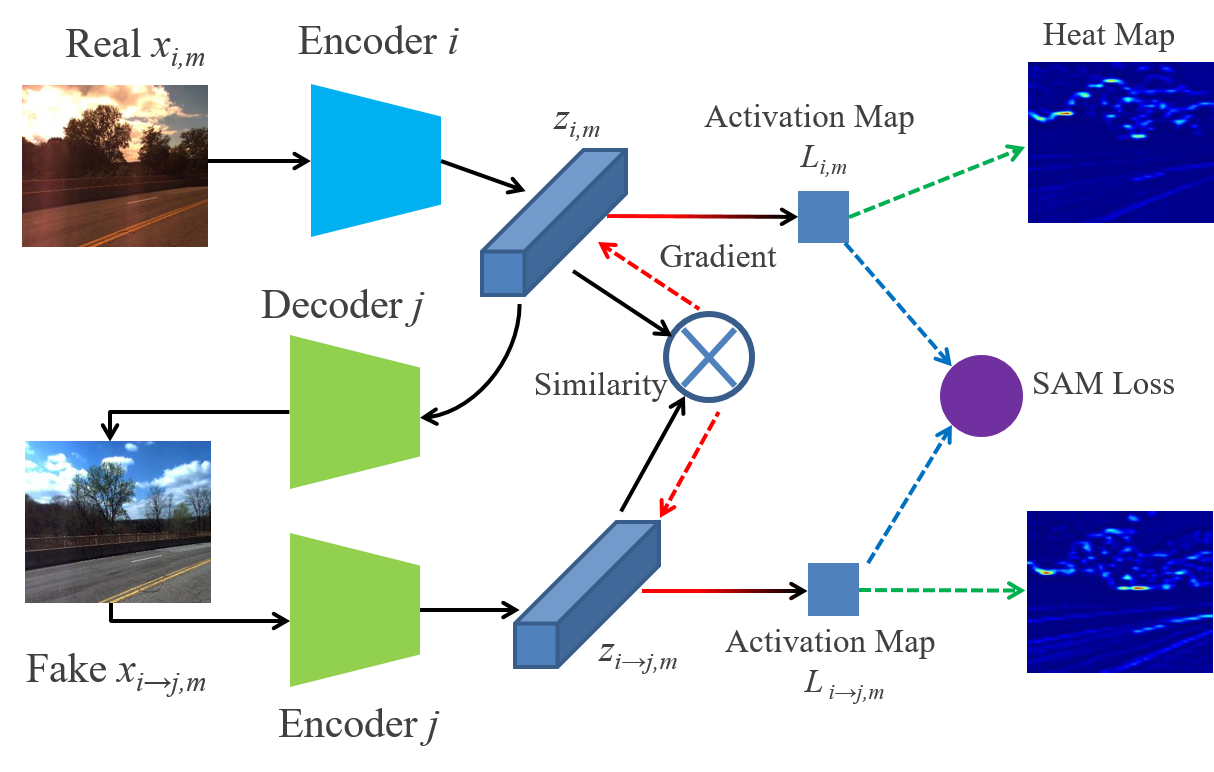}
	%	\vspace{-0.4cm}
	%}
	%\includegraphics[scale=1.0]{figurefile}
	\caption
	{
		The illustration of one branch of SAM loss from domain $ i $ to $ j $. The real image in domain $ i $ is first translated to fake image in domain $ j $, and  the gradient of similarity w.r.t each other could be calculated, denoted as red dashed lines. And then the activation map is the sum of feature map weighted by the gradient, shown as color-gradient line from red to black, and SAM loss could be calculated in a self-supervised manner. \textcolor{ColorName}{Note that the notation of $ L_{i,m}$ and $ L_{i\to j,m} $ here are short for  $ L_{i,j,m} $ and  $ L_{i\to j,i,m} $ derived from Equation (\ref{equ17}) and (\ref{equ18}).}
	}
	\label{SAM_fig}
	%	\vspace{-0.4cm}
\end{figure}

\subsection{Feature Consistency Loss}
% exlain the loss 
% prove it works
To obtain the shared-latent feature across different domains, different from the way in which \cite{liu2017unsupervised} does, \textcolor{ColorName}{an additional loss exerted on the latent space, called feature consistency loss, is proposed in \cite{hu2019retrieval}.}  Under the above assumptions, for image $ x_{i} $ from domain $ i $ it is formulated as,
\begin{eqnarray}
\label{fcl_loss}    
\mathcal{L}_{FCL}^{x_{i}}=\mathbb{E}_{x_{i} \sim p(x_{i})}[\|E_{j}^{c}((D_{j}^{c}(E_{i}^{c}(x_{i}))))-E_{i}^{c}(x_{i})\|_{2}]
\end{eqnarray}
As a result, the domain-invariant feature \cite{hu2019retrieval} could be extracted by combining all the weighted losses together,
\begin{eqnarray}
\label{DIFL_loss}    
\mathcal{L}_{DIF}=  \mathcal{L}_{GAN}^{x_{i}} + \mathcal{L}_{GAN}^{x_{j}} + \lambda_{cyc}(\mathcal{L}_{Cycle}^{x_{i}}+\mathcal{L}_{Cycle}^{x_{j}}) \notag \\ 
+\lambda_{FCL}(\mathcal{L}_{FCL}^{x_{i}}+\mathcal{L}_{FCL}^{x_{j}}) 
\end{eqnarray}

Here gives the theoretical analysis for FCL. Supposing the optimality of the DIF loss (\ref{DIFL_loss}) is reached, Equation (\ref{equ4}) to (\ref{equ11}) are still satisfied. Additionally, because of the feature consistency loss (\ref{fcl_loss}), based on Equation (\ref{equ4}) (\ref{equ6}) (\ref{equ10})we have,
\begin{eqnarray}
\label{equ14}    
z_{i\to j,m} = z_{i,m} \\
\label{equ15}  
E_{i}^{c} = (D_{i}^{c})^{-1} 
\end{eqnarray}

Since $ z_{i\to j,m} $ and $ i $ are independent which is discussed in the last section, $ z_{i,m} $ and $ i $ are independent for any domain $ i $ from Equation (\ref{equ14}), which indicates that the latent feature is well-shared across multiple domains and represents the content latent code given any image from any domain. Furthermore, the trained encoders and decoders are inverse and the goal of finding underlying encoders $ E_{i}^{c*} $ and decoders $ D_{i}^{c*} $ is reached according to Section \ref{sec3sub1} . So it is appropriate to use the content latent code for image representation across different environmental conditions.

\section{Coarse-to-fine Retrieval-based Localization}
\label{sec4}

\subsection{Gradient-weighted Similarity Activation Mapping Loss}
However the original domain-invariant feature (\ref{DIFL_loss}) cannot excavate the context or localizing information of the content latent feature map, as a result the performance of place recognition under high accuracy is limited. To this end, we propose a novel gradient-weighted similarity activation mapping loss for shared-latent feature to fully discover the weighted similar area for high-accuracy retrieval.

Inspired by CAM\cite{zhou2016learning}, Grad-CAM \cite{selvaraju2017grad} and Grad-CAM++ \cite{chattopadhay2018grad} in visual explanation for classification with convolutional neural networks, we assume that the place recognition task could be regarded as an extension of image multi-classification with infinite target classes, where each database image represents a single target class for each query image during the retrieval process. Then for each query image, the similarity to each database image is treated as the score before softmax or probability for multi-classification task and the one with the largest similarity is the retrieved result, which is similar to the classification result with the largest probability.

Ideally, suppose the identical content latent feature maps from domain $ i, j $, $ z_{i,m},z_{j,m} $, have the shape of $ n \times h\times w $, where identical content $ m $ is omitted for brevity. First the mean value of the cosine similarity on the height and width dimension is calculated below,
\begin{eqnarray}
\label{equ16}
Y =\frac{1}{n} \sum_{k} \cdot \frac{ \sum_{p} \sum_{q} z^{kpq}_{i} z^{kpq}_{j}}{\sqrt{\sum_{p,q} {z^{kpq}_{i}}^{2}}\sqrt{\sum_{p,q} {z^{kpq}_{j}}^{2}}} 
\end{eqnarray}
where $ z^{kpq}_{i} $ represents the $ k $th, $ p $th, and $ q $th on the dimension of channel, height and width, i.e. $ n \times h\times w $ for the content feature map $ z_{i} $.

$ Y $ is the score of similarity between $ z_{i}$ and $z_{j} $. Following the definition of Grad-CAM \cite{selvaraju2017grad}, we have the similarity activation weight and map,
\textcolor{ColorName}{
\begin{eqnarray}
\label{equ17}
\omega^{k}_{i,j} =  \sum_{p}\sum_{q} \frac{\partial Y}{\partial z^{kpq}_{i}}
\end{eqnarray}
\begin{eqnarray}
\label{equ18}
L^{pq}_{i,j} = ReLU(\sum_{k}\omega^{k}_{i,j}z^{kpq}_{i})
\end{eqnarray}
where $ \omega^{k}_{i,j} $ is the weight that $ z_{j}^{k} $ gives to $ z_{i}^{k} $ and $ L^{pq}_{i,j}  $ is the activation map of $ z_{i}^{k} $ at the position $ (p,q) $ given the database reference of $ z_{j}^{k} $. $ z_{j} $ is treated as database feature ("class label") for the query feature $ z_{i} $ in  the place recognition task. And if we take partial derivative \textit{w.r.t} $ z^{kpq}_{j} $ in Equation  (\hyperref[equ17]{\ref{equ17}}), the weight $ \omega^{k}_{j,i} $ and the activation map $ L^{pq}_{j,i} $ would result in that $ z_{i} $ is the retrieval from database under domain $ i $ given query feature $  z_{j} $. $ L^{pq}_{i,j} $ is unequal to $ L^{pq}_{j,i} $ due to the different weight in Equation (\hyperref[equ17]{\ref{equ17}}). We denote $ L_{i,j} $ as $ L_i $ and $ L_{j,i} $ as $ L_j $ for short in the following notation.}

Equation (\hyperref[equ17]{\ref{equ17}}) and Equation (\hyperref[equ17]{\ref{equ18}}) are the mathematic formulation of the proposed gradient-weighted similarity activation mapping (Grad-SAM), where activation map  is aggregated by each gradient-weighted feature map, retaining the localizing information of deep feature map.  In order to only input the positively-activated areas for training, we exert a ReLU function to obtain the final activation map $ L_{i,j} $ or  $ L_{j,i} $.

\begin{figure}[thpb]
	\centering
	%\framebox
	%{
	%\begin{mdframed}[hidealllines=true,backgroundcolor=blue!20]
		%\lipsum[2]
	\textcolor{ColorName}{\includegraphics[scale=0.43]{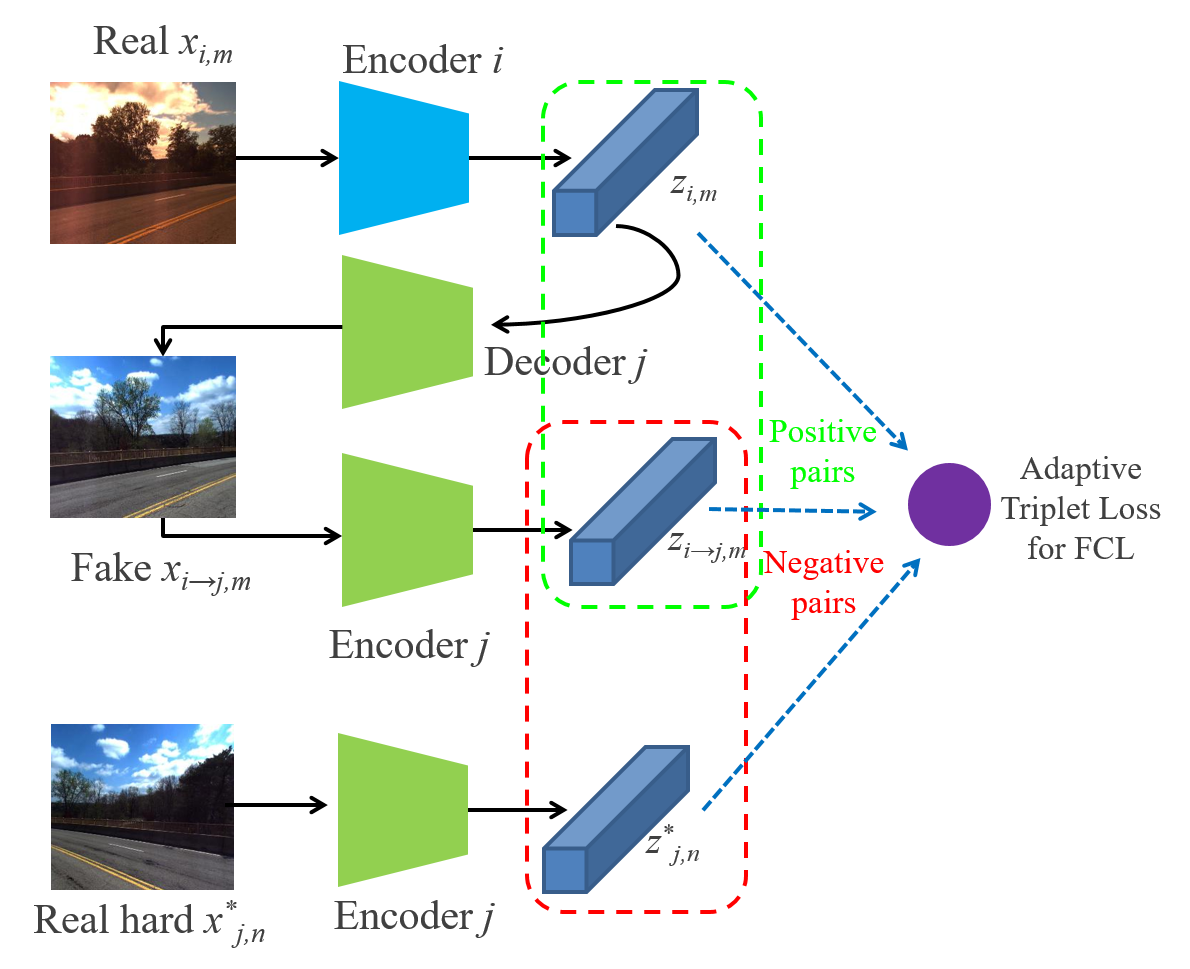}}
	%\end{mdframed}
	
	%	\vspace{-0.4cm}
	%}
	%\includegraphics[scale=1.0]{figurefile}
	\caption
	{
		The illustration of one-branch adaptive triplet loss for FCL from domain $ i $ to $ j $. The inputs of the loss are the encoded latent features from real images in domain $ i,j $ and the translated image $ i\to j $, resulting in the negative pairs with the red dashed box and the positive pairs with the green dashed box. \textcolor{ColorName}{Note that positive pairs only differ in the environment while the place is the only difference for the negative pairs.}
	}
	\label{FCL_trip_fig}
	%	\vspace{-0.4cm}
\end{figure}

Particularly, as shown in Figure \ref{SAM_fig} , inside the unsupervised DIFL architecture, the content latent codes $ z_{i,m}, z_{j,n} $ are shared from the same distribution but $ z_{i,m} \neq z_{j,n} $ for the unpaired $ m \neq n $. The similarity activation map $ L_{i,m}, L_{i\to j,m} $ could be visualized by resizing to the original size in Figure \ref{SAM_fig}. \textcolor{ColorName}{According to FCL loss (\ref{fcl_loss}), $ z_{i,m} $ and $ z_{i \to j, m} $ tend to be identical, which means that the calculation of similarity between them is meaningful and so is the SAM loss. Therefore, the self-supervised Grad-SAM loss for domain $ i $ could be formulated below based on Equation (\ref{equ16} - \ref{equ18}).
\begin{eqnarray}  
\label{SAM_loss}   
\mathcal{L}_{SAM}^{x_i}=\mathbb{E}_{x_{i} \sim p(x_{i})}[\|L_{i,m}-L_{i\to j,m}\|_{2}   ]
\end{eqnarray}
where $ z_{i,m} $ and $ z_{i \to j, m} $ are substituted into $ z_{i} $ and $ z_{j} $ in Equation (\ref{equ16} - \ref{equ18}) and $ L_{i,m}$ and $ L_{i\to j,m} $ are short for  $ L_{i,j,m} $ and  $ L_{i\to j,i,m} $ derived from Equation (\ref{equ17}) and (\ref{equ18}).}

\subsection{Adaptive Triplet Loss}
% problem: scene is not specific in the latent space, all scene distance is small...
% no labels and no pairs for triplet networks 
%formula, how to adapt and keep balance, representation first , metric later
Though the domain-invariant feature learning is obtained through feature consistency loss (\ref{fcl_loss}) and Grad-SAM loss (\ref{SAM_loss}) is for further finer retrieval with salient localizing information on the latent feature map, it is difficult to distinguish different latent content codes using domain-invariant feature without explicit metric learning. As the  distance of the latent features with the same content is decreasing due to feature consistency loss (\ref{fcl_loss}) and Grad-SAM loss (\ref{SAM_loss}), the distance of latent features for different contents may be forced to diminish as well, resulting in mismatched retrievals for test images in long-term visual localization.

\begin{figure}[thpb]
	\centering
	%\framebox
	%{
	%\begin{mdframed}[hidealllines=true,backgroundcolor=blue!20]
		%\lipsum[2]
	\includegraphics[scale=0.36]{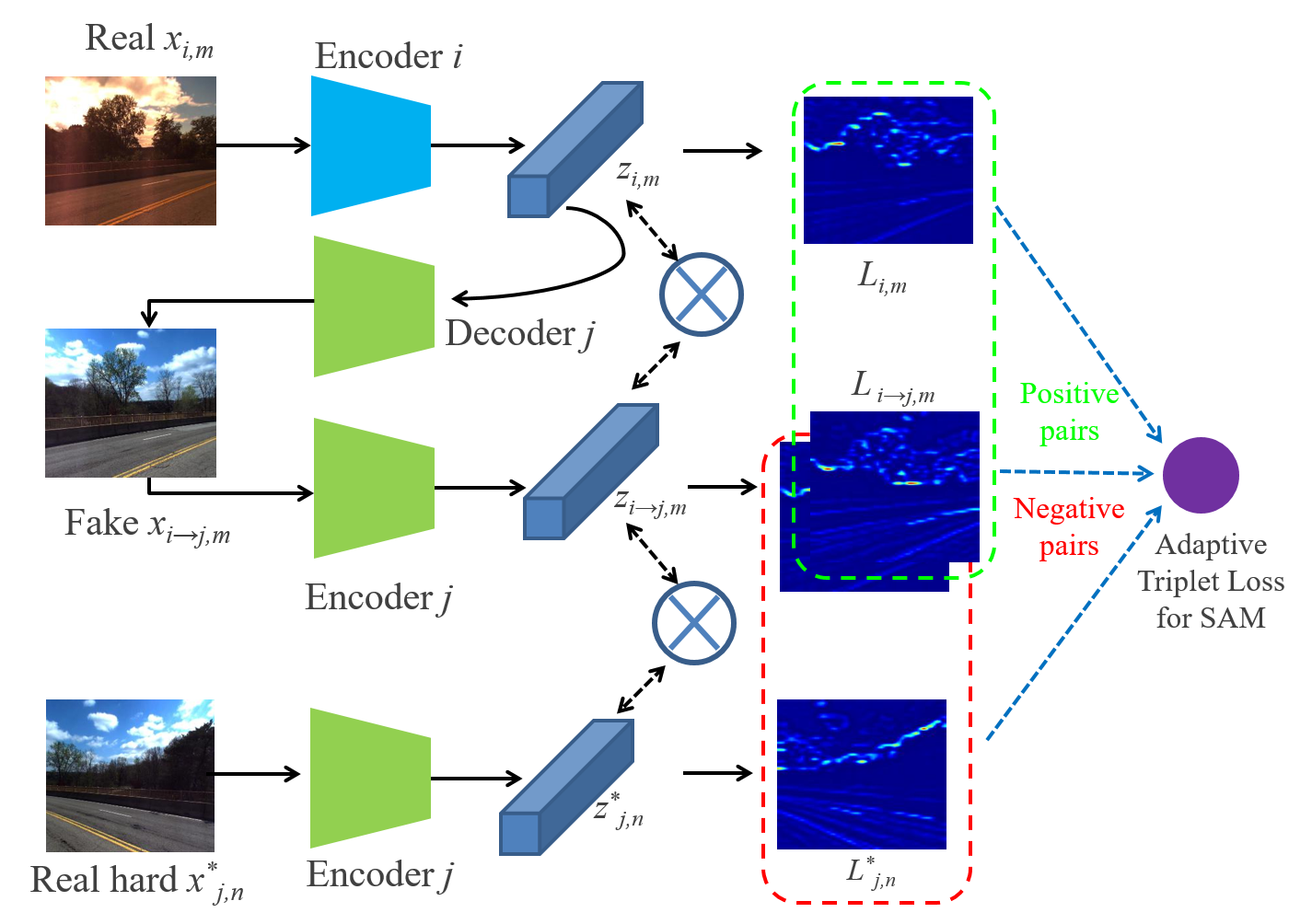}
	%\end{mdframed}
	
	%	\vspace{-0.4cm}
	%}
	%\includegraphics[scale=1.0]{figurefile}
	\caption
	{
		The illustration of one-branch adaptive triplet loss for Grad-SAM from domain $ i $ to $ j $. The inputs of the loss are the similarity activation maps from real images in domain $ i,j $ and the translated image $ i\to j $. The negative pairs are bounded with the red dashed box while the positive pairs are bounded with the green dashed box. Note that the activation maps in domain $ i $ from two pairs are slightly different.
	}
	\label{SAM_trip_fig}
	%	\vspace{-0.4cm}
\end{figure}

\label{why_flip_and_hard}
Toward this end, we propose a novel adaptive triplet loss based on  feature consistency loss (\ref{fcl_loss}) and Grad-SAM loss (\ref{SAM_loss}) to improve the contrastive learning of the latent representation inside the self-supervised DIFL framework. Suppose unpaired images $ x_{i,m}, x_{j,n} $ are selected from domain $ i, j, i\neq j $,  where $ m,n $ represent the content of images. \textcolor{ColorName}{Note that for the purpose of unsupervised training pipeline, one of the selected image is horizontally flipped while the other is not so that $ m \neq n $ is assured for the negative pair. The operation of flipping only one of the input images is random and also functions as data augmentation due to the fact that the flipped images follow the distribution of original images. Details could be found in Section \ref{sec4sub1} .} For the self-supervised contrastive learning, the positively paired samples are not given but generated from the framework in Equation (\ref{equ4} - \ref{equ6}) and (\ref{equ16} - \ref{equ18}), \textit{i.e.} $ z_{i,m}, z_{i\to j,m} $ and $ L_{i,m}, L_{i\to j,m} $. \textcolor{ColorName}{For the negatively paired samples, for the sake of the fact that the images under the same environmental condition tend to be closer than ones under different conditions do, the stricter constraint is implemented for negative pairs with the translated image and the other real image, which are under the same environment but different places, \textit{i.e.} $ z_{i\to j,m}, z_{j,n} $ and $ L_{i\to j,m}, L_{j,n} $.}

\textcolor{ColorName}{Moreover, in order to improve the efficiency of the triplet loss for representation learning during the late iterations, the negative pair with the least distance between the original and the translated one is automatically selected as the hard negative pair $ z^*_{j,n} $ or $ L^*_{j,n} $ from a group of random negative candidates $ \mathbb{Z}_{j,n} $ or $  \mathbb{L}_{j,n} $, shown as Equation (\ref{hard_negative1} , \ref{hard_negative2}). The adaptive triplet loss is calculated through these hard negative pairs without any supervision or extra priority information.
	\begin{eqnarray}  
	\label{hard_negative1}  
	z^*_{j,n} = \mathop{\arg\min}_{z_{j,n} \in \mathbb{Z}_{j,n} }\|z_{i\to j,m} - z_{j,n}\|_2 \\
	\label{hard_negative2}
	L^*_{j,n} = \mathop{\arg\min}_{L_{j,n} \in \mathbb{L}_{j,n}}\|L_{i\to j,m} - L_{j,n}\|_2
	\end{eqnarray}
}

We adopt the basic form of triplet loss from \cite{wohlhart2015learning}, but the \textit{margin} depends on the feature consistency loss (\ref{fcl_loss}) or Grad-SAM loss (\ref{SAM_loss}), which adapts to the representation learning of (\ref{fcl_loss}) or (\ref{SAM_loss}). The illustrations of the adaptive triplet loss for FCL and SAM are shown in Figure \ref{FCL_trip_fig} and \ref{SAM_trip_fig} .  The adaptive triplet loss for FCL and Grad-SAM for domain $ i $ is shown below,
\textcolor{ColorName}{
\begin{eqnarray}  
\label{fcl_triplet_loss}   
&&\mathcal{L}_{Triplet\_FCL}^{x_i}=\mathbb{E}_{x_{i} \sim p(x_{i})}[      \max (0, 1- \\ 
&& \frac{\|z_{i\to j,m} - z^*_{j,n}\|_{2}}{\|z_{i,m} - z_{i\to j,m}\|_{2} + m_{f}\exp(-\alpha_{f}\|z_{i,m} - z_{i\to j,m} \|_{2} )})]\notag
\end{eqnarray}
\begin{eqnarray}  
\label{SAM_triplet_loss}   
&&\mathcal{L}_{Triplet\_SAM}^{x_i}=\mathbb{E}_{x_{i} \sim p(x_{i})}[      \max (0,1- \\ 
&&  \frac{\|L_{i\to j,m} - L^*_{j,n}\|_{2}}{\|L_{i,m} - L_{i\to j,m}\|_{2} + m_{s}\exp(-\alpha_{s}\|L_{i,m} - L_{i\to j,m} \|_{2} )})]\notag
\end{eqnarray}
}
where hyperparameters $ m_{f}, m_s $ are the \textit{margin}, which is the value that the distance of negative pairs exceeds the distance of self-generated positive pairs when the image translation is well trained, \textit{i.e.} $ p(x_{i\to j,m})= p(x_{j,m}) $. However constant \textit{margin} has an influence on the joint model training with FCL or Grad-SAM loss, so we propose the self-adaptive term, which is the exponent function of negative FCL loss or Grad-SAM loss weighted by $ \alpha_{f} $ or $ \alpha_{s} $.

Combining with the adaptive triplet loss (\ref{fcl_triplet_loss}) or (\ref{SAM_triplet_loss}), in the beginning of the whole model training, the exponential adaptive term is close to 0 so the triplet loss term does not affect the FCL (\ref{fcl_loss}) or Grad-SAM (\ref{SAM_loss}). But as the training process goes by, the triplet loss would dominate the model training since the  exponential adaptive term becomes larger and closer to 1.

\subsection{Coarse-to-fine Image Retrieval Pipeline}
\label{coarse2fine}
% training: siam+trip, siam+sam+samtrip, total loss
% how to test, how to implement, database and query, top 3
\textcolor{ColorName}{For the image retrieval, we adopt the coarse-to-fine strategy to fully leverage the models with different training settings for different specific purposes. The DIFL model with FCL (\ref{fcl_loss}) and triplet loss  (\ref{fcl_triplet_loss}) aims to find the database retrieval for the query image using general domain-invariant features and results in better performance of localization within larger error threshold, shown in the Section \ref{ablation_study} , which gives a good initial range of retrieved candidates and could be used as a coarse retrieval. 
The total loss for coarse-retrieval model training is shown below,
\begin{eqnarray}
\label{coarse_loss}    
\mathcal{L}_{coarse}=  \mathcal{L}_{GAN}^{x_{i}} + \mathcal{L}_{GAN}^{x_{j}} + \lambda_{cyc}(\mathcal{L}_{Cycle}^{x_{i}}+\mathcal{L}_{Cycle}^{x_{j}})  \\ 
+\lambda_{FCL}(\mathcal{L}_{FCL}^{x_{i}}+\mathcal{L}_{FCL}^{x_{j}}) 
+ \notag \\  \lambda_{Triplet\_FCL}(\mathcal{L}_{Triplet\_FCL}^{x_i}+\mathcal{L}_{Triplet\_FCL}^{x_j})\notag
\end{eqnarray}
where $ \lambda_{cyc}, \lambda_{FCL}, \lambda_{Triplet\_FCL} $ are the hyperparameters to weigh different loss terms.}
	
\textcolor{ColorName}{Furthermore, to obtain the finer retrieval results, we incorporate the Grad-SAM (\ref{SAM_loss}) with its triplet loss (\ref{SAM_triplet_loss}) into the coarse-retrieval model, which fully digs out the localizing information of feature map and promotes the high-accuracy retrieval across different conditions shown in Table \ref{ablation}. However, according to Section \ref{ablation_study} , the accuracy of low-precision localization for fine-retrieval model is lower than the coarse-retrieval model, which shows the necessity of the coarse retrieval at first. The total loss for the finer model training is shown below,
\begin{table}[h]
	\caption{Condition Correspondence for RoborCar Dataset}
	\label{Correspondence}
	\begin{center}
		\begin{tabular}{|c|c|c|c|}
			\hline
			\multicolumn{2}{|c|}{\begin{tabular}[c]{@{}c@{}}\textbf{Conditions}\\  \textbf{in RobotCar}\end{tabular}} & \multicolumn{2}{c|}{\begin{tabular}[c]{@{}c@{}}\textbf{the Most Similar} \\ \textbf{Conditions in CMU-Seasons}\end{tabular}} \\ \hline
			Description                                      & Date                                 & Description                                              & Date                                 \\ \hline
			Overcast(reference)                               & 28 Nov                               & Overcast, Mixed Foliage                                  & 28 Oct                               \\ \hline
			Dawn                                             & 16 Dec                               & Low Sun, Mixed Foliage                                   & 12 Nov                               \\ \hline
			Dusk                                             & 20 Feb                               & Low Sun, Foliage                                         & 4 Mar                                \\ \hline
			Night                                            & 10 Dec                               & Overcast, Mixed Foliage                                  & 28 Oct                               \\ \hline
			Night-rain                                       & 17 Dec                               & Overcast, Mixed Foliage                                  & 28 Oct                               \\ \hline
			Overcast-summer                                  & 22 May                               & Overcast, Foliage                                        & 28 Jul                               \\ \hline
			Overcast-winter                                  & 13 Nov                               & Cloudy, Foliage                                          & 1 Oct                                \\ \hline
			Rain                                             & 25 Nov                               & Cloudy, Mixed Foliage                                    & 22 Nov                               \\ \hline
			Snow                                             & 3 Feb                                & Low Sun, Snow                                & 21 Dec                               \\ \hline
			Sun                                              & 10 Mar                               & Sunny                                     & 4 Apr                                \\ \hline
		\end{tabular}
	\end{center}
\end{table}
\begin{eqnarray}
\label{fine_loss}    
\mathcal{L}_{fine}=  \mathcal{L}_{GAN}^{x_{i}} + \mathcal{L}_{GAN}^{x_{j}} + \lambda_{cyc}(\mathcal{L}_{Cycle}^{x_{i}}+\mathcal{L}_{Cycle}^{x_{j}})  \\ 
+\lambda_{FCL}(\mathcal{L}_{FCL}^{x_{i}}+\mathcal{L}_{FCL}^{x_{j}}) 
+ \lambda_{SAM}(\mathcal{L}_{SAM}^{x_i}+\mathcal{L}_{SAM}^{x_j}) \notag \\  +\lambda_{Triplet\_SAM}(\mathcal{L}_{Triplet\_SAM}^{x_i}+\mathcal{L}_{Triplet\_FCL}^{x_j})\notag \\
+\lambda_{Triplet\_FCL}(\mathcal{L}_{Triplet\_FCL}^{x_i}+\mathcal{L}_{Triplet\_FCL}^{x_j})\notag
\end{eqnarray}
where $ \lambda_{cyc}, \lambda_{FCL},\lambda_{SAM},  \lambda_{Triplet\_SAM}, \lambda_{Triplet\_FCL} $ are the hyperparameters for each loss term.}

Once the coarse and fine models are trained, the test pipeline contains coarse retrieval and finer retrieval. The 6-DoF poses of database images are given while the goal is to find the poses of query images.  We first pre-encode each database image under the reference environment into feature map through coarse model off line, forming the database of coarse features. While testing, for every query image, we extract the feature map using coarse encoder of the corresponding domain and retrieve the \textit{top-N}  most similar ones from pre-encoded coarse features in the database. The $ N $ candidates are then encoded through the fine model to find the secondary feature maps, and the query image is also encoded through the fine model to find the query feature. The most similar one in the  $ N $ candidates is retrieved as the final result for localization. \textcolor{ColorName}{Although the coarse-to-fine strategy may not get the most similar retrieval globally in some cases, it will increase the accuracy within coarse error in Section \ref{ablation_study} compared to the only single fine model, which is beneficial to the application of pose regression for relocalization. Also it may benefit from the filtered coarse candidates in some cases in Table \ref{ablation} to improve medium-precision results.} The 6-DoF pose of query image is the same as the finally-retrieved one in the database.

\section{Experimental Results}
\label{sec5}

We conduct a series of experiments on CMU-Seasons dataset and validate the effectiveness of coarse-to-fine pipelines with the proposed FCL loss, Grad-SAM loss and adaptive triplet loss. With the model only trained  on the urban part of CMU seasons dataset in an unsupervised manner, we compare our results with several image-based localization baselines on the untrained suburban and park part of CMU-Seasons dataset and RobotCar-Seasons dataset, showing the advantage under scenes with massive vegetation and robustness to huge illumination change. We conduct these experiments on two NVIDIA 2080Ti cards with 64G RAM on Ubuntu 18.04 system. \textcolor{ColorName}{Our source code and pre-trained models are available on \href{https://github.com/HanjiangHu/DISAM}{\texttt{https://github.com/HanjiangHu/DISAM}}.}

\begin{table}[h]
	\caption{Results Comparison to Baselines on CMU-Seasons Dataset}
	%\vspace{-0.4cm}
	\label{result_cmu_total}
	\begin{center}
		\begin{tabular}{c|c|c|c}
			\hline
			\multirow{3}{*}{\textbf{Method}} & \textbf{Park(\%)}                                                                         & \textbf{Suburban(\%)}                                                                     & \textbf{Urban(\%)}                                                                        \\
			& 0.25m / 0.5m                                                                              & 0.25m / 0.5m                                                                              & 0.25m / 0.5m                                                                              \\
			& 2$^{\circ}$ / 5$^{\circ}$ & 2$^{\circ}$ / 5$^{\circ}$ & 2$^{\circ}$ / 5$^{\circ}$ \\ \hline
			FAB-MAP\cite{cummins2008fab}                          & 0.8 / 1.7                                                                                 & 0.5 / 1.5                                                                                 & 2.7 / 6.4                                                                                 \\
			NetVLAD\cite{arandjelovic2017netvlad}                          & 5.6 / 15.7                                                                                & 7.7 / 21.0                                                                                & 17.4 / 40.3                                                                               \\
			DenseVLAD\cite{torii201824}                        & 10.3 / 27.1                                                                               & 9.8 / 26.6                                                                                & 22.2 / \textbf{48.6}                                                                               \\
			DIFL-FCL\cite{hu2019retrieval}                         & 11.4 / 28.9                                                                               & 9.7 / 25.0                                                                                & 20.2 / 44.7                                                                               \\ \hline
			\textcolor{ColorName}{\textbf{Coarse-only(ours)}}                           & \textcolor{ColorName}{11.3} / \textcolor{ColorName}{29.1}                                                                              & \textcolor{ColorName}{9.9} / \textcolor{ColorName}{25.6}                                                                                & \textcolor{ColorName}{20.1} / \textcolor{ColorName}{45.0}                                                                               \\
			\textcolor{ColorName}{\textbf{Fine-only(ours)}}                             & \textcolor{ColorName}{\textbf{13.2}} / \textcolor{ColorName}{\textbf{32.2}}                                                                               & \textcolor{ColorName}{\textbf{11.3}} / \textcolor{ColorName}{27.2}                                                                               & \textcolor{ColorName}{\textbf{22.7}} / \textcolor{ColorName}{46.4}                                                                               \\
			\textcolor{ColorName}{\textbf{Coarse2Fine(ours)}}                      & \textcolor{ColorName}{12.6} / \textcolor{ColorName}{31.3}                                                                               & \textcolor{ColorName}{11.1} / \textcolor{ColorName}{\textbf{27.5}}                                                                               & \textcolor{ColorName}{22.6} / \textcolor{ColorName}{47.3}                                                                               \\ \hline
		\end{tabular}
	\end{center}
\end{table}

\begin{table}[h]
	\caption{ Comparison with Baselines on Foliage Condition \protect \\
		\textcolor{ColorName}{Reference is No Foliage}}
	%\vspace{-0.4cm}
	\label{result_cmu_fol}
	\begin{center}
		\begin{tabular}{c|c|c}
			\hline
			\multirow{3}{*}{Method} & \textbf{Foliage(\%)}      & \textbf{Mixed Foliage(\%)} \\
			& 0.25m / 0.5m              & 0.25m / 0.5m               \\
			& 2$^{\circ}$ / 5$^{\circ}$ & 2$^{\circ}$ / 5$^{\circ}$  \\ \hline
			FAB-MAP\cite{cummins2008fab}                 & 1.1 / 2.7                 & 1.0 / 2.5                  \\
			NetVLAD\cite{arandjelovic2017netvlad}                 & 10.4 / 26.1               & 11.0 / 26.7                \\
			DenseVLAD\cite{torii201824}               & 13.2 / 31.6               & 16.2 / 38.1                \\
			DIFL-FCL\cite{hu2019retrieval}                & 13.9 / 32.7               & 16.6 / 38.6                \\ \hline
			\textcolor{ColorName}{\textbf{Coarse-only(ours)}}           & \textcolor{ColorName}{14.0} / \textcolor{ColorName}{33.1}               & \textcolor{ColorName}{16.4} / \textcolor{ColorName}{38.5}                \\
			\textcolor{ColorName}{\textbf{Fine-only(ours)}}           & \textcolor{ColorName}{\textbf{15.3}} / \textcolor{ColorName}{33.5}               & \textcolor{ColorName}{\textbf{19.1}} / \textcolor{ColorName}{\textbf{42.2}}                \\
			\textcolor{ColorName}{\textbf{Coarse2Fine(ours)}}           & \textcolor{ColorName}{15.2} / \textcolor{ColorName}{\textbf{34.1}}               & \textcolor{ColorName}{18.7} / \textcolor{ColorName}{41.7}                \\ \hline
		\end{tabular}
	\end{center}
	%\vspace{-0.5cm}
\end{table}

\subsection{Experimental Setup} 
\label{sec4sub1}

The first series of experiments are conducted on the CMU-Seasons dataset\cite{sattler2018benchmarking}, which is derived from the CMU Visual Localization \cite{Badino2011} dataset. It was recorded by a vehicle with a left-side and a right-side camera  over a year along the route about 9 kilometers in Pittsburgh, U.S. The environmental change of seasons, illumination and especially foliage is very challenging on this dataset. \cite{sattler2018benchmarking} benchmarks the dataset and presents the  groudtruth of camera pose only for the reference database images, adding new categories and area divisions of the original dataset as well. There are 31250 images in 7 slices for urban area,13736 images in 3 slices for suburban area and 30349 images in 7 slices for park area. Each area part has only one reference and eleven query environmental conditions. The condition of database is \textit{Sunny + No Foliage}, and conditions of query images could be any weather intersected with vegetation condition, e.g. \textit{Overcast + Mixed Foliage}. \textcolor{ColorName}{Since the images in training dataset contain both the left-side and right-side ones, the operation of flipping horizontally is reasonable and acceptable for the unsupervised generation of negative pairs and data augmentation, as introduced in Section \ref{why_flip_and_hard} .}

The second series of experiments are conducted on RobotCar Seasons dataset \cite{sattler2018benchmarking} derived from Oxford RobotCar dataset \cite{RobotCarDatasetIJRR}. The images were captured with three Point Grey Grasshopper2 cameras on the left, rear and right of the vehicle along the 10km route under changing weather, season and illumination across a year in Oxford, U.K. It contains 6954 triplets for database images under overcast condition, 3100 triplets for day-time query images under 7 conditions and 878 triplets for night-time images under 2 conditions. In the experiment we only test rear images with the pre-trained model on the urban part of CMU-Seasons dataset to validate the generalization ability of our approach. Considering that not all the conditions of RoborCar datasets have exactly corresponding conditions in CMU-Seasons, we choose the pre-trained models under the conditions with the most similar descriptions and dates from CMU-Seasons dataset for all the conditions in RobotCar dataset listed in Table \ref{Correspondence} . Note that for the conditions which are not included in CMU-Seasons, we use the pre-trained models under the reference condition instead, \textit{Overcast + Mixed Foliage}, for the sake of  fairness.

The images are scaled to $ 286 \times 286$ and cropped to $ 256 \times 256$ size randomly while training but directly scaled to $ 256 \times 256$ while testing, leading to the shape of feature map with the shape of  $ 256\times64\times64 $. We follow  the protocol introduced in \cite{sattler2018benchmarking} which is the percentage of correctly-localized query images. \textcolor{ColorName}{Since we only focus on the high and medium precision, the pose error thresholds are $ (0.25m, 2^{\circ})$ and $(0.5m, 5^{\circ})$ while coarse-precision (low-precision) $(5m, 10^{\circ}) $ is omitted for the purpose of high-precision localization except for the ablation study.} We choose several image-based localization methods FAB-MAP\cite{cummins2008fab}, DIFL-FCL \cite{hu2019retrieval}, NetVLAD\cite{arandjelovic2017netvlad} and DenseVLAD \cite{torii201824}, which are the best image-based localization methods.

\subsection{Evaluation on CMU-Seasons Dataset}%change of result of base 1200 and 600 compare with fcl 600,ablation study (l2 or cos lambda2 or pca 1*256*72*96)
\label{sec4sub2}

\begin{figure*}[htbp]
	
	\centering
	
	\subfigure[]{
		\begin{minipage}[t]{0.095\linewidth}
			\centering
			\includegraphics[width=0.7in]{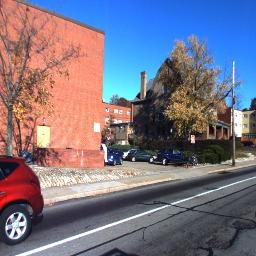}\\
%			\vspace{0.1cm}
			\includegraphics[width=0.7in]{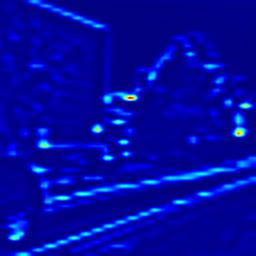}\\
%			\vspace{0.1cm}
			%\caption{fig1}
		\end{minipage}%
	\begin{minipage}[t]{0.095\linewidth}
		\centering
		\includegraphics[width=0.7in]{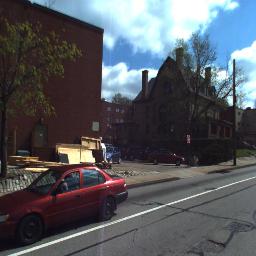}\\
%		\vspace{0.1cm}
		\includegraphics[width=0.7in]{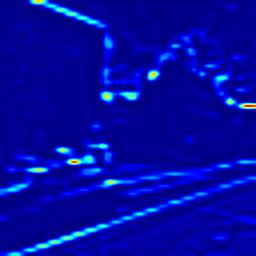}\\
%		\vspace{0.1cm}
		%\caption{fig1}
	\end{minipage}%
	}%
	\subfigure[]{
			\begin{minipage}[t]{0.095\linewidth}
				\centering
				\includegraphics[width=0.7in]{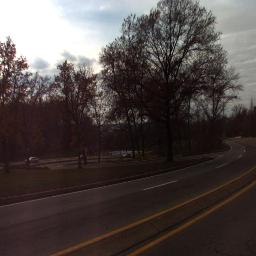}\\
				%			\vspace{0.1cm}
				\includegraphics[width=0.7in]{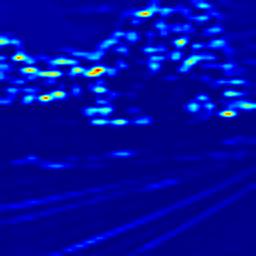}\\
				%			\vspace{0.1cm}
				%\caption{fig1}
			\end{minipage}%
			\begin{minipage}[t]{0.095\linewidth}
				\centering
				\includegraphics[width=0.7in]{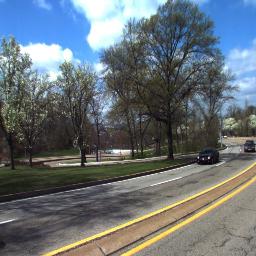}\\
				%		\vspace{0.1cm}
				\includegraphics[width=0.7in]{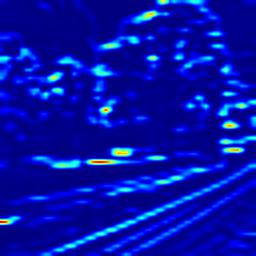}\\
				%		\vspace{0.1cm}
				%\caption{fig1}
			\end{minipage}%
		}%
		\subfigure[]{
		\begin{minipage}[t]{0.095\linewidth}
			\centering
			\includegraphics[width=0.7in]{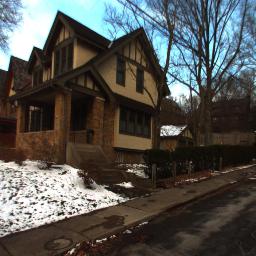}\\
			%			\vspace{0.1cm}
			\includegraphics[width=0.7in]{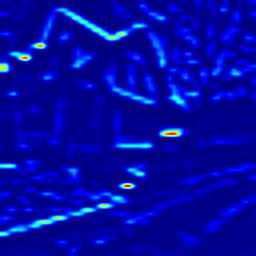}\\
			%			\vspace{0.1cm}
			%\caption{fig1}
		\end{minipage}%
		\begin{minipage}[t]{0.095\linewidth}
			\centering
			\includegraphics[width=0.7in]{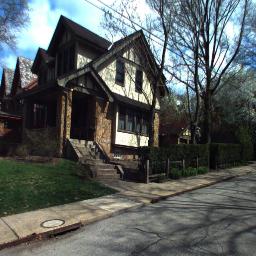}\\
			%		\vspace{0.1cm}
			\includegraphics[width=0.7in]{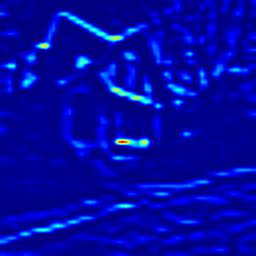}\\
			%		\vspace{0.1cm}
			%\caption{fig1}
		\end{minipage}%
	}%
	\subfigure[]{
	\begin{minipage}[t]{0.095\linewidth}
		\centering
		\includegraphics[width=0.7in]{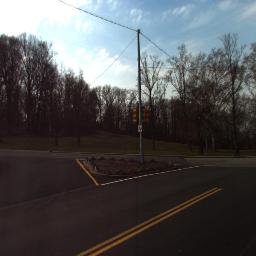}\\
		%			\vspace{0.1cm}
		\includegraphics[width=0.7in]{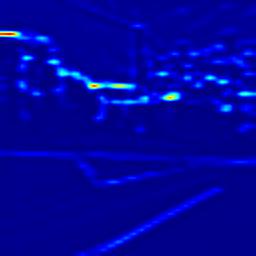}\\
		%			\vspace{0.1cm}
		%\caption{fig1}
	\end{minipage}%
	\begin{minipage}[t]{0.095\linewidth}
		\centering
		\includegraphics[width=0.7in]{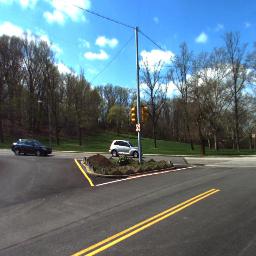}\\
		%		\vspace{0.1cm}
		\includegraphics[width=0.7in]{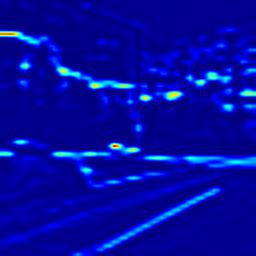}\\
		%		\vspace{0.1cm}
		%\caption{fig1}
	\end{minipage}%
}%
	\subfigure[]{
	\begin{minipage}[t]{0.095\linewidth}
		\centering
		\includegraphics[width=0.7in]{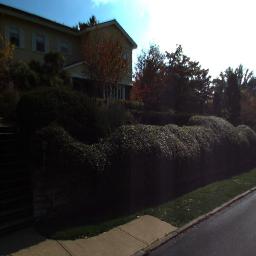}\\
		%			\vspace{0.1cm}
		\includegraphics[width=0.7in]{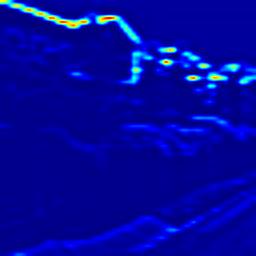}\\
		%			\vspace{0.1cm}
		%\caption{fig1}
	\end{minipage}%
	\begin{minipage}[t]{0.095\linewidth}
		\centering
		\includegraphics[width=0.7in]{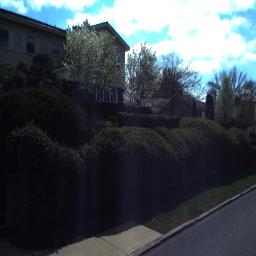}\\
		%		\vspace{0.1cm}
		\includegraphics[width=0.7in]{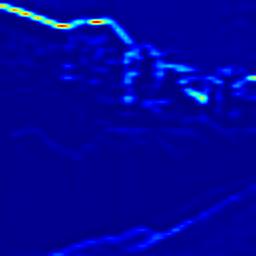}\\
		%		\vspace{0.1cm}
		%\caption{fig1}
	\end{minipage}%
}%

	\centering
	%	\vspace{-0.2cm}
	\caption{Results on CMU-Seasons dataset. For each set of images in (a) to (e), the top left is the query image while the top right is the database image under the condition of \textit{Sunny + No Foliage}. The query images of Set (a) to (e) are under the conditions of \textit{Low Sun + Mixed Foliage}, \textit{Overcast + Mixed Foliage}, \textit{Low Sun + Snow}, \textit{Low Sun + Foliage} and \textit{Sunny + Foliage} respectively. The visualizations of similarity activation maps are on the bottom row for all the query or database RGB images. }
	%	\vspace{-0.5cm}
	\label{cmu_results}
\end{figure*}

 Following the transfer learning strategy for DIFL in \cite{hu2019retrieval}, \textcolor{ColorName}{we fine-tune the pre-trained models in \cite{hu2019retrieval} at epoch 300 which are trained only with cycle consistency loss and GAN loss under all the images of CMU-Seasons Dataset for pure image translation task. Then for the representation learning task, the model is fine-tuned with images of Urban area in an unsupervised manner, without paired images across conditions.} After adding the other loss terms in (\ref{coarse_loss}) or (\ref{fine_loss}), we continue to train until epoch 600 with learning rate linearly decreasing from  0.0002 to 0. Then the model is trained in the same manner until epoch 1200 with split 300 epochs. \textcolor{ColorName}{In order to speed up and stabilize the training process with triplet loss, we use the random negative pairs from epoch 300 to epoch 600 for the fundamental representation learning and adopt the hard negative pairs from epoch 600, as shown in Section \ref{why_flip_and_hard} . We choose the hard negative pair from 10 pairs of negative samples for each iteration.}

For the coarse-retrieval model training, the weight hyperparameter are maximumly set as  $ \lambda_{cyc}=10, \lambda_{FCL}=0.1, \lambda_{Triplet\_FCL}=1 $, which are all linearly increasing from 0 as the training process goes by to balance the multi-task framework. \textcolor{ColorName}{Similarly for the fine-retrieval model training, we set $ \lambda_{cyc}=10, \lambda_{FCL}=0.1, \lambda_{SAM}=1000,  \lambda_{Triplet\_SAM}=1, \lambda_{Triplet\_FCL}=1 $ with the similar training strategy. The fine model consists of the metrics of both $ L2 $ and $ cosine \ similarity $ for FCL terms while only $ L2 $ metric is used in the coarse model for FCL terms.} For the adaptive triplet loss, we set $ m_{f}=5, \alpha_{f}=2$ in triplet FCL loss (\ref{fcl_triplet_loss}) and $ m_{s}=0.1, \alpha_{s}=1000$ in triplet SAM loss (\ref{SAM_triplet_loss}). 
\begin{table}[htbp]
	\caption{Comparison with Baselines on Weather Condition \protect \\
		\textcolor{ColorName}{Reference is Sunny} }
	%\vspace{-0.4cm}
	\label{result_cmu_weatherl}
	\begin{center}
		\begin{tabular}{c|c|c|c}
			\hline
			\multirow{3}{*}{\textbf{Method}} & \textbf{Overcast(\%)} & \textcolor{ColorName}{\textbf{Cloudy(\%)}} & \textbf{Low Sun(\%)} \\
			& 0.25m / 0.5m       & 0.25m / 0.5m           & 0.25m / 0.5m         \\
			& 2$^{\circ}$ / 5$^{\circ}$            & 2$^{\circ}$ / 5$^{\circ}$                & 2$^{\circ}$ / 5$^{\circ}$               \\ \hline
			FAB-MAP\cite{cummins2008fab}                 & 0.9 / 2.7           & \textcolor{ColorName}{1.8} / \textcolor{ColorName}{4.1}       & 2.0 / 4.6          \\
			NetVLAD\cite{arandjelovic2017netvlad}                 & 10.9 / 27.0        & \textcolor{ColorName}{13.0} / \textcolor{ColorName}{30.5}     & 10.1 / 25.7       \\
			DenseVLAD\cite{torii201824}               & 15.1 / 35.2        & \textcolor{ColorName}{18.4} / \textcolor{ColorName}{\textbf{41.8}}    & 15.1 / 36.9       \\
			DIFL-FCL\cite{hu2019retrieval}                & 15.9 / 36.9       & \textcolor{ColorName}{16.4} / \textcolor{ColorName}{37.6}     & 13.9 / 34.1       \\ \hline
			\textcolor{ColorName}{\textbf{Coarse-only(ours)}}           & \textcolor{ColorName}{15.9} / \textcolor{ColorName}{36.9}        & \textcolor{ColorName}{16.5} / \textcolor{ColorName}{38.2}     & \textcolor{ColorName}{13.9} / \textcolor{ColorName}{34.4}       \\
			\textcolor{ColorName}{\textbf{Fine-only(ours)}}           & \textcolor{ColorName}{\textbf{18.3}} / \textcolor{ColorName}{39.4}        & \textcolor{ColorName}{\textbf{18.9}} / \textcolor{ColorName}{40.3}     & \textcolor{ColorName}{ \textbf{16.2}} / \textcolor{ColorName}{\textbf{37.7}}       \\
			\textcolor{ColorName}{\textbf{Coarse2Fine(ours)}}           & \textcolor{ColorName}{18.0} / \textcolor{ColorName}{\textbf{39.6}}        & \textcolor{ColorName}{18.6} / \textcolor{ColorName}{40.5}     & \textcolor{ColorName}{15.8} / \textcolor{ColorName}{37.3}       \\ \hline
		\end{tabular}
	\end{center}
	%\vspace{-0.5cm}
\end{table}
\begin{table}[htbp]
	\caption{Results Comparison to Baselines on RobotCar Dataset }
	%\vspace{-0.4cm}
	\label{result_robotcar}
	\begin{center}
		\begin{tabular}{c|c|c}
			\hline
			\multirow{3}{*}{\textbf{Method}} & \textbf{Night all(\%)}     & \textbf{Day all(\%)}       \\
			& 0.25m / 0.5m        & 0.25m / 0.5m        \\
			& 2$^{\circ}$ / 5$^{\circ}$  & 2$^{\circ}$ / 5$^{\circ}$ \\ \hline
			FAB-MAP\cite{cummins2008fab}                          & 0.0 /  0.0             & 2.7 / 11.8             \\
			NetVLAD\cite{arandjelovic2017netvlad}                          & 0.3 / 2.3            & 6.4 / 26.3            \\
			DenseVLAD\cite{torii201824}                        & 1.0 / 4.4           & 7.6 / \textbf{31.2}          \\
			DIFL+FCL\cite{hu2019retrieval}                         & 2.5 / 6.5            & 7.6 / 26.2           \\ \hline
			\textcolor{ColorName}{\textbf{Coarse-only(ours)}}                    & \textcolor{ColorName}{\textbf{3.3}} / \textcolor{ColorName}{\textbf{8.7}}            & \textcolor{ColorName}{7.4} / \textcolor{ColorName}{25.8}          \\ 
			\textcolor{ColorName}{\textbf{Fine-only(ours)}}                    & \textcolor{ColorName}{2.1} / \textcolor{ColorName}{4.8}            & \textcolor{ColorName}{\textbf{8.2}} / \textcolor{ColorName}{27.0}          \\ \hline
		\end{tabular}
	\end{center}
	%\vspace{-0.3cm}
\end{table}
 And during the two-stage retrieval, the number of coarse candidates \textit{ top-N}  is set to be $ 3 $, which makes it both efficient and effective. In the two-stage retrieval pipeline, we use the mean value of the $ cosine \ similarity $ on the height and width dimension as the metric during the coarse retrieval, as shown in Equation (\ref{equ16}). And for the fine retrieval, we use the normal $ cosine\ similarity $ for the flatten secondary features due to the salient information in the feature map.

Our final result is compared with baselines shown as Table  \ref{result_cmu_total} , where ours outperforms baseline methods for high- and medium-precision localization, $(0.25m, 2^{\circ})$ and $ (0.5m, 5^{\circ}) $, in park and suburban area, which shows powerful generalization ability because the model is only trained on the urban area.  The medium-precision localization in the urban area is affected by numerous dynamic objects. 

 We further compare the performance on different foliage categories from \cite{sattler2018benchmarking}, \textit{Foliage} and \textit{ Mixed Foliage} \textcolor{ColorName}{with the reference database under \textit{No Foliage}}, which is the most challenging problem for this dataset. The results are shown in Table \ref{result_cmu_fol}  , from which we can see that our result is better than baselines under different conditions of foliage for the localization with medium and high precision. \textcolor{ColorName}{To investigate the performance under different weather conditions, we compare the models with baselines on the \textit{Overcast}, \textit{Cloudy} and \textit{Low Sun} conditions with the reference database under \textit{Sunny} in Table \ref{result_cmu_weatherl}, which covers almost all the weather conditions.  It could be seen that our results present the best medium- and high-accuracy  results on most of the weather conditions. The \textit{Cloudy} weather contains plenty of cloud in the sky, which provides some noise on the activation map for fine retrieval with reference to the clear sky under \textit{Sunny}, which could be regarded as a kind of dynamic objects.}
 
 \textcolor{ColorName}{From the results of different areas, vegetation and weather, it could be seen that the finer retrieval boosts the results of coarse retrieval. Moreover, the coarse-to-fine retrieval strategy gives better performance than the fine-only method in some cases, showing the significance and effectiveness for high- and medium-precision localization of the two-stage strategy.} The reasonable explanation for the good performance under different foliage and weather conditions lies in that the latent content code is robust and invariant for changing vegetation and illumination.  All the results (including ours) are from the official benchmark website of long-term visual localization \cite{sattler2018benchmarking}. Some results of fine-retrieval are shown in Figure \ref{cmu_results}, where the activation maps give the localizing information of feature maps and the salient areas mostly exist around the edges or adjacent parts of different instance patches due to the gradient-based activation.

\subsection{Evaluation on RobotCar Dataset}%baselines no densevlad
\label{sec4sub3}

\begin{figure*}[htbp]
	
	\centering
	
	\subfigure[]{
		\begin{minipage}[t]{0.095\linewidth}
			\centering
			\includegraphics[width=0.7in]{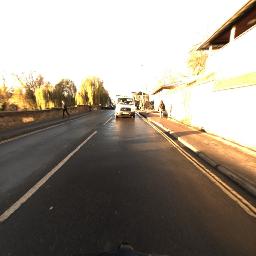}\\
			%			\vspace{0.1cm}
			\includegraphics[width=0.7in]{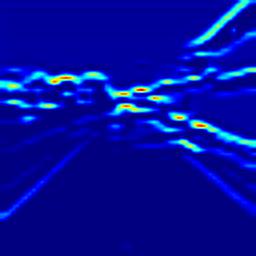}\\
			%			\vspace{0.1cm}
			%\caption{fig1}
		\end{minipage}%
		\begin{minipage}[t]{0.095\linewidth}
			\centering
			\includegraphics[width=0.7in]{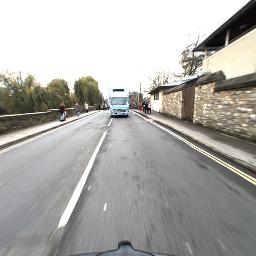}\\
			%		\vspace{0.1cm}
			\includegraphics[width=0.7in]{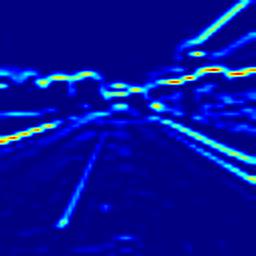}\\
			%		\vspace{0.1cm}
			%\caption{fig1}
		\end{minipage}%
	}%
	\subfigure[]{
		\begin{minipage}[t]{0.095\linewidth}
			\centering
			\includegraphics[width=0.7in]{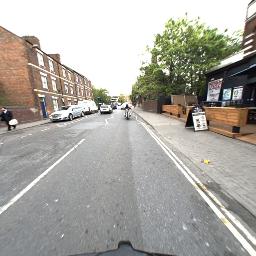}\\
			%			\vspace{0.1cm}
			\includegraphics[width=0.7in]{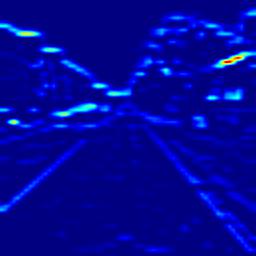}\\
			%			\vspace{0.1cm}
			%\caption{fig1}
		\end{minipage}%
		\begin{minipage}[t]{0.095\linewidth}
			\centering
			\includegraphics[width=0.7in]{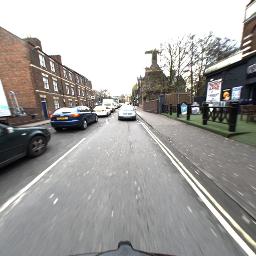}\\
			%		\vspace{0.1cm}
			\includegraphics[width=0.7in]{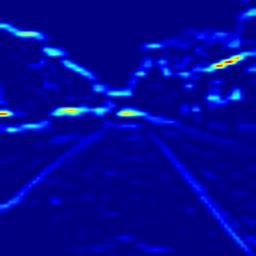}\\
			%		\vspace{0.1cm}
			%\caption{fig1}
		\end{minipage}%
	}%
\subfigure[]{
	\begin{minipage}[t]{0.095\linewidth}
		\centering
		\includegraphics[width=0.7in]{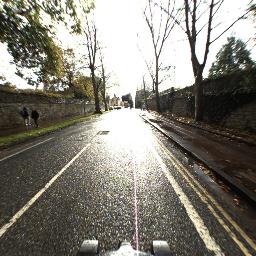}\\
		%			\vspace{0.1cm}
		\includegraphics[width=0.7in]{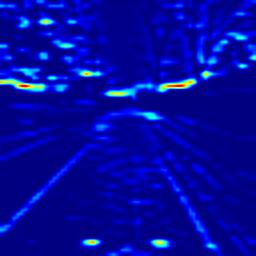}\\
		%			\vspace{0.1cm}
		%\caption{fig1}
	\end{minipage}%
	\begin{minipage}[t]{0.095\linewidth}
		\centering
		\includegraphics[width=0.7in]{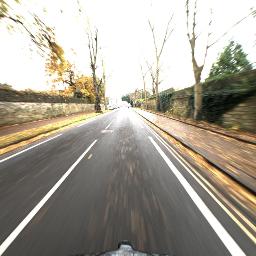}\\
		%		\vspace{0.1cm}
		\includegraphics[width=0.7in]{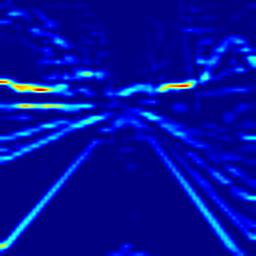}\\
		%		\vspace{0.1cm}
		%\caption{fig1}
	\end{minipage}%
}%
\subfigure[]{
	\begin{minipage}[t]{0.095\linewidth}
		\centering
		\includegraphics[width=0.7in]{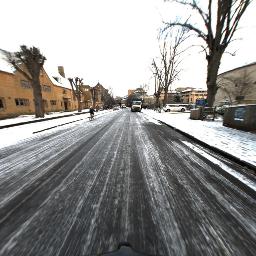}\\
		%			\vspace{0.1cm}
		\includegraphics[width=0.7in]{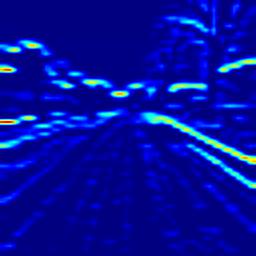}\\
		%			\vspace{0.1cm}
		%\caption{fig1}
	\end{minipage}%
	\begin{minipage}[t]{0.095\linewidth}
		\centering
		\includegraphics[width=0.7in]{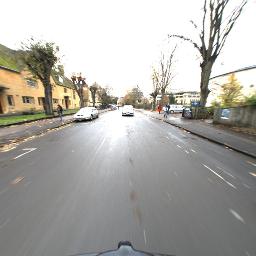}\\
		%		\vspace{0.1cm}
		\includegraphics[width=0.7in]{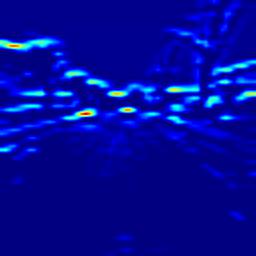}\\
		%		\vspace{0.1cm}
		%\caption{fig1}
	\end{minipage}%
}%
\subfigure[]{
	\begin{minipage}[t]{0.095\linewidth}
		\centering
		\includegraphics[width=0.7in]{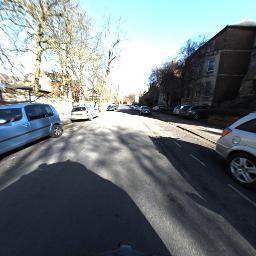}\\
		%			\vspace{0.1cm}
		\includegraphics[width=0.7in]{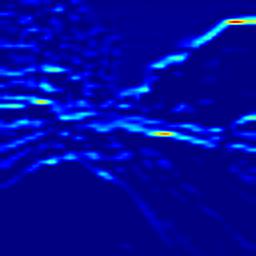}\\
		%			\vspace{0.1cm}
		%\caption{fig1}
	\end{minipage}%
	\begin{minipage}[t]{0.095\linewidth}
		\centering
		\includegraphics[width=0.7in]{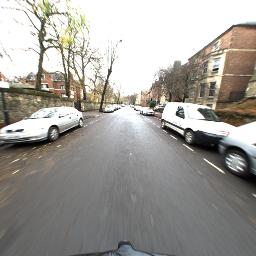}\\
		%		\vspace{0.1cm}
		\includegraphics[width=0.7in]{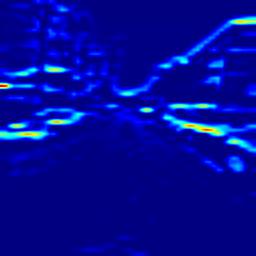}\\
		%		\vspace{0.1cm}
		%\caption{fig1}
	\end{minipage}%
}%
	\centering
	%	\vspace{-0.2cm}
	\caption{Results on RobotCar dataset. For each set of images in (a) to (e), the top left is the day-time query image while the top right is the database image under the condition of \textit{Overcast}. The query images of Set (a) to (e) are under the conditions of \textit{Dawn}, \textit{Overcast-summer}, \textit{Overcast-winter}, \textit{Snow} and \textit{Sun} respectively.  The visualizations of similarity activation maps are on the bottom row for all the query or database RGB images. }
	%	\vspace{-0.5cm}
	\label{robotcar_results}
\end{figure*}

\begin{table*}[htbp]
	\caption{Ablation Study on Different Strategies and Loss Terms}
	%\vspace{-0.4cm}
	\label{ablation}
	\begin{center}
	\begin{tabular}{|c|c|c|c|c|c|c|c|}
		\hline
		&                                &                                        &                                &                                        & \textbf{Urban(\%)}                                & \textbf{Suburban(\%)}                            & \textbf{Park(\%)}                                 \\
		&                                &                                        &                                &                                        & 0.25m / 0.5m / \textcolor{ColorName}{5m}                        & 0.25m / 0.5m / \textcolor{ColorName}{5m}                        & 0.25m / 0.5m / \textcolor{ColorName}{5m}                        \\
		\multirow{-3}{*}{\textbf{Strategy}} & \multirow{-3}{*}{\textbf{FCL}} & \multirow{-3}{*}{\textbf{Triplet FCL}} & \multirow{-3}{*}{\textbf{SAM}} & \multirow{-3}{*}{\textbf{Triplet SAM}} & 2$^{\circ}$ / 5$^{\circ}$ / \textcolor{ColorName}{10$^{\circ}$} & 2$^{\circ}$ / 5$^{\circ}$ / \textcolor{ColorName}{10$^{\circ}$} & 2$^{\circ}$ / 5$^{\circ}$ / \textcolor{ColorName}{10$^{\circ}$} \\ \hline
		& $\times$                       & $\times$                               & $\times$                       & $\times$                               & \textcolor{ColorName}{19.8} / \textcolor{ColorName}{44.1} / \textcolor{ColorName}{85.8}                       & \textcolor{ColorName}{9.0} / \textcolor{ColorName}{23.2} / \textcolor{ColorName}{66.9}                        & \textcolor{ColorName}{10.0} / \textcolor{ColorName}{26.3} / \textcolor{ColorName}{69.6}                       \\
		& $\checkmark$                   & $\times$                               & $\times$                       & $\times$                               & \textcolor{ColorName}{19.8} / \textcolor{ColorName}{44.3} / \textcolor{ColorName}{86.8}                       & \textcolor{ColorName}{9.2} / \textcolor{ColorName}{23.5} / \textcolor{ColorName}{69.4}                        & \textcolor{ColorName}{10.1} / \textcolor{ColorName}{26.5} / \textcolor{ColorName}{71.6}                        \\
		& $\checkmark$                   & Constant Margin                        & $\times$                       & $\times$                               & \textcolor{ColorName}{20.0} / \textcolor{ColorName}{44.7} / \textcolor{ColorName}{88.5}  & \textcolor{ColorName}{9.7} / \textcolor{ColorName}{25.3} / \textcolor{ColorName}{77.5}   & \textcolor{ColorName}{11.2} / \textcolor{ColorName}{28.6} / \textcolor{ColorName}{78.3}  \\
		\multirow{-4}{*}{Coarse-only}       & $\checkmark$                   & Adaptive Margin                        & $\times$                       & $\times$                               & \textcolor{ColorName}{20.1} / \textcolor{ColorName}{45.0} / \textcolor{ColorName}{\textbf{89.9}}  & \textcolor{ColorName}{9.9} / \textcolor{ColorName}{25.6} / \textcolor{ColorName}{\textbf{78.6}}   & \textcolor{ColorName}{11.3} / \textcolor{ColorName}{29.1} / \textcolor{ColorName}{\textbf{80.1}}  \\ \hline
		& $\checkmark$                   & $\times$                               & $\checkmark$                   & $\times$                               & \textcolor{ColorName}{21.2} / \textcolor{ColorName}{44.8} / \textcolor{ColorName}{80.6}                       & \textcolor{ColorName}{10.0} / \textcolor{ColorName}{24.1} / \textcolor{ColorName}{65.8}                       & \textcolor{ColorName}{11.1} / \textcolor{ColorName}{27.9} / \textcolor{ColorName}{68.3}                       \\
		& $\checkmark$                   & $\times$                               & $\checkmark$                   & Constant Margin                        & \textcolor{ColorName}{22.3} / \textcolor{ColorName}{45.5} / \textcolor{ColorName}{83.7}  & \textcolor{ColorName}{11.0} / \textcolor{ColorName}{26.0} / \textcolor{ColorName}{68.4}  & \textcolor{ColorName}{12.9} / \textcolor{ColorName}{30.9} / \textcolor{ColorName}{73.2}  \\
		& $\checkmark$                   & $\times$                               & $\checkmark$                   & Adaptive Margin                        & \textcolor{ColorName}{22.5} / \textcolor{ColorName}{46.0} / \textcolor{ColorName}{84.6}                       & \textcolor{ColorName}{11.1} / \textcolor{ColorName}{26.9} / \textcolor{ColorName}{69.8}                       & \textcolor{ColorName}{13.0} / \textcolor{ColorName}{31.5} / \textcolor{ColorName}{74.7}                       \\
		\multirow{-4}{*}{Fine-only}         & \textcolor{ColorName}{$\checkmark$}                   & \textcolor{ColorName}{Adaptive Margin}                        & \textcolor{ColorName}{$\checkmark$}                   & \textcolor{ColorName}{Adaptive Margin}                        & \textcolor{ColorName}{\textbf{22.7}} / \textcolor{ColorName}{46.4} / \textcolor{ColorName}{85.4}  & \textcolor{ColorName}{\textbf{11.3}} / \textcolor{ColorName}{27.2} / \textcolor{ColorName}{71.9}  & \textcolor{ColorName}{\textbf{13.2}} / \textcolor{ColorName}{\textbf{32.2}} / \textcolor{ColorName}{76.9}  \\ \hline
		Coarse-to-fine                      & $\checkmark$                   & Adaptive Margin                        & $\checkmark$                   & Adaptive Margin                        & \textcolor{ColorName}{22.6} / \textcolor{ColorName}{\textbf{47.3}} / \textcolor{ColorName}{89.1}  & \textcolor{ColorName}{11.1} / \textcolor{ColorName}{\textbf{27.5}} / \textcolor{ColorName}{77.6}  & \textcolor{ColorName}{12.6} / \textcolor{ColorName}{31.3} / \textcolor{ColorName}{80.0}  \\ \hline
	\end{tabular}
\end{center}
\end{table*}

In order to further validate the generalization ability of our proposed method to the unseen scenarios, we directly use the pre-trained models on urban area of CMU-Seasons to test on the RobotCar dataset, according to the correspondent condition from CMU-Seasons for every query condition of RobotCar based on Table \ref{Correspondence} . \textcolor{ColorName}{Considering the database images are much more than query images under each condition, the two-stage strategy is skipped for practicality and efficiency, only testing coarse-only and fine-only models. The metric for both coarse and fine retrieval is the mean value of the $ cosine \ similarity $ on the height and width dimension as shown in Equation (\ref{equ16}).}

The comparison results are shown in Table \ref{result_robotcar} , where ours outperforms other baseline methods under the \textit{Night} and \textit{Night-rain} conditions. Note that the model we use for the night-time retrieval is the same as the database  because night-time images are not included in the training set, showing the effectiveness of the representation learning in the latent space form autoencoder-structured model. \textcolor{ColorName}{Since the images under \textit{Night} and \textit{Night-rain} conditions have too poor context or localizing information to find the correct similarity activation maps, the coarse model performs better than the finer model.}

 Our results under all the \textit{Day} conditions are the best for high-precision performance, showing the powerful generalization ability in the unknown scenarios and environments through attaining satisfactory retrieval-based localization results.  All the results (including ours) are also from the official benchmark website of long-term visual localization \cite{sattler2018benchmarking}. Some day-time results are shown in Figure \ref{robotcar_results}, including all the environments which have similar ones among pre-trained models on CMU-Seasons dataset. 

\subsection{Ablation Study}
\label{ablation_study}
For the further ablation study in Table \ref{ablation}, we implement different strategies (\textit{Coarse-only}, \textit{Fine-only} and \textit{Coarse-to-fine}) and different loss terms (\textit{FCL}, \textit{Triplet FCL}, \textit{SAM}, and \textit{Triplet SAM}) during model training, and test them on CMU-Seasons dataset. \textcolor{ColorName}{The only difference between \textit{Coarse-only} and \textit{Fine-only} lies in whether the model is trained with \textit{SAM} or not, while coarse-to-fine strategy follows the two-stage strategy in Section \ref{coarse2fine} . It could be seen that  \textit{Coarse-only} models perform the best in low-precision localization, which is suitable to provide the rough candidates for the upcoming finer retrieval. With the incorporation of \textit{SAM}-related loss, the medium- and high-precision accuracy increase while the low-precision one decreases. The \textit{Coarse-to-fine} combines the advantages of \textit{Coarse-only} and \textit{Fine-only} together, improving the the low-precision localization of fine models as well as the medium- and high-precision localization of coarse models simultaneously, which shows the effectiveness and significance of the two-stage strategy by overcoming both the weaknesses. Furthermore, because of the high-quality potential candidates provided by \textit{Coarse-only} model, some medium-precision results of \textit{Coarse-to-fine} on the last row perform the best and other results are extremely close the best ones, which shows the promising performance of the two-stage strategy.}

\textcolor{ColorName}{From the first two rows of Table \ref{ablation}, the DIFL with FCL performs better than vanilla ComboGAN (\ref{combogan_loss}), which indicates that FCL assists to extract the domain-invariant feature. Due to the effective self-supervised triplet loss with hard negative pairs, the performance with \textit{Triplet FCL} or \textit{Triplet SAM} is significantly improved compared with the results on the second or fifth row respectively. To validate the effectiveness of \textit{Adaptive Margin} in triplet loss, we compare the results of \textit{Constant Margin} and \textit{Adaptive Margin}, which show that the model with adaptive margin gives better results than that with constant margin for both \textit{Triplet FCL} and \textit{Triplet SAM}. The last row in \textit{Fine-only} strategy shows the hybrid adaptive triplet losses of both FCL and SAM are beneficial to the fine retrieval. Note that the settings of training and testing for Table \ref{ablation} are consistent internally, but are slightly different from the experimental settings in the  \cite{hu2019retrieval} in many aspects, like training epochs, the metrics for retrieval, the choice of the pre-trained models for testing, \textit{etc.}.}

\section{Conclusion}
\label{sec6}

In this work, we have formulated the domain-invariant feature learning architecture for long-term retrieval-based localization with feature consistency loss (FCL). Then a novel loss based on gradient-weighted similarity activation map (Grad-SAM) is proposed for the improvement of high-precision performance. The adaptive triplet loss based on FCL loss or Grad-SAM loss is incorporated to the framework to form the coarse or fine retrieval, resulting in the coarse-to-fine testing pipeline. Our proposed method is also compared with several state-of-the-art image-based localization baselines on CMU-Seasons and RobotCar-Seasons dataset, where our results outperform the baseline methods for image retrieval in medium- and high-precision localization in challenging environments. However, there are a few concerns about our method that the performance under the dynamic scenes is week compared to other image-based methods, which could be addressed by adding semantic information to enhance the robustness to dynamic objects in the future. 

\ifCLASSOPTIONcaptionsoff
  \newpage
\fi



%\bibliographystyle{IEEEtran}
%\begin{thebibliography}{1}


%\end{thebibliography}
\bibliographystyle{IEEEtran}
\bibliography{IEEEabrv,mylib}
%\vspace{-1cm}
\begin{IEEEbiography}[{\includegraphics[width=1in,height=1.25in,clip,keepaspectratio]{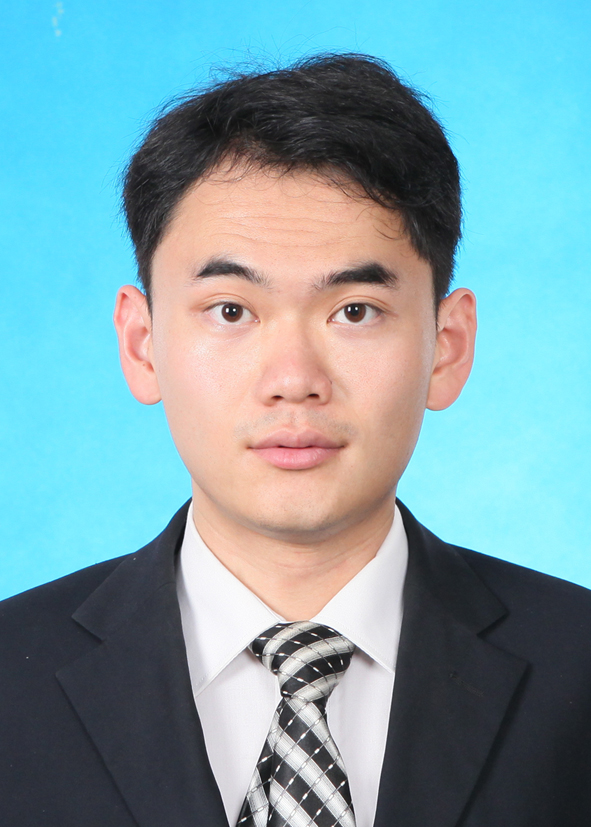}}]{Hanjiang Hu}
	received the B.Eng degree in Mechanical Engineering from Shanghai Jiao Tong University, Shanghai, China, in 2018, and M.S. degree in Control Science and Engineering from Shanghai Jiao Tong University, Shanghai, China, in 2021. He is currently working toward  the Ph.D. degree in Mechanical Engineering and secondary M.S. degree in Machine Learning at Carnegie Mellon University, Pittsburgh, PA, USA. His current research interests include robust perception, multi-agent system, and machine learning.
\end{IEEEbiography}
%\vspace{-3cm}
\begin{IEEEbiography}[{\includegraphics[width=1in,height=1.25in,clip,keepaspectratio]{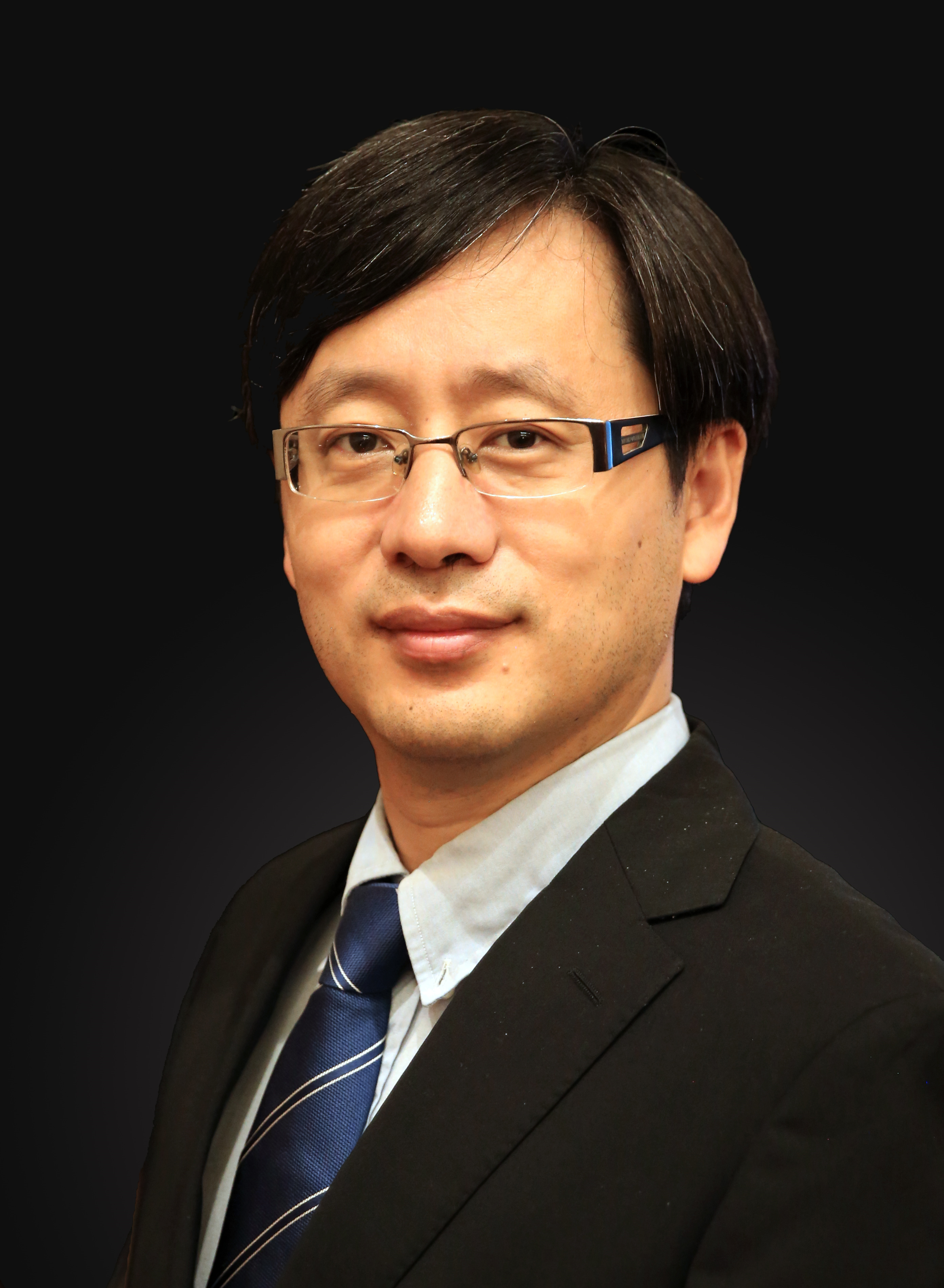}}]{Hesheng Wang}
	(SM’15) received the B.Eng. degree in electrical engineering from the Harbin Institute of Technology, Harbin, China, in 2002, and the M.Phil. and Ph.D. degrees in automation and computer-aided engineering from The Chinese University of Hong Kong, Hong Kong, in 2004 and 2007, respectively. He was a Post-Doctoral Fellow and Research Assistant with the Department of Mechanical and Automation Engineering, The Chinese University of Hong Kong, from 2007 to 2009. He is currently a Professor with the Department of Automation, Shanghai Jiao Tong University, Shanghai, China. His current research interests include visual servoing, service robot, adaptive robot control, and autonomous driving. 
	Dr. Wang is an Associate Editor of Assembly Automation and the International Journal of Humanoid Robotics, a Technical Editor of the IEEE/ASME TRANSACTIONS ON MECHATRONICS. He served as an Associate Editor of the IEEE TRANSACTIONS ON ROBOTICS from 2015 to 2019. He was the General Chair of the IEEE RCAR 2016, and the Program Chair of the IEEE ROBIO 2014 and IEEE/ASME AIM 2019.
\end{IEEEbiography}
\vspace{-3cm}
\begin{IEEEbiography}[{\includegraphics[width=1in,height=1.25in,clip,keepaspectratio]{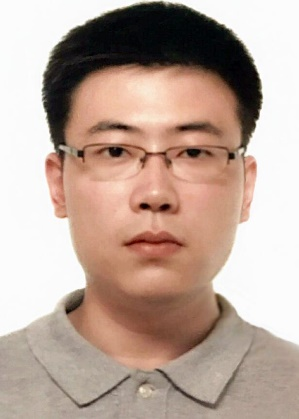}}]{Zhe Liu}
	received his B.S. degree in Automation from Tianjin University, Tianjin, China, in 2010, and Ph.D. degree in Control Technology and Control Engineering from Shanghai Jiao Tong University, Shanghai, China, in 2016. From 2017 to 2020, he was a Post-Doctoral Fellow with the Department of Mechanical and Automation Engineering, The Chinese University of Hong Kong, Hong Kong. He is currently a Research Associate with the Department of Computer Science and Technology, University of Cambridge. His research interests include autonomous mobile robot, multirobot cooperation and autonomous driving system.
\end{IEEEbiography}
\vspace{-3cm}
\begin{IEEEbiography}[{\includegraphics[width=1in,height=1.25in,clip,keepaspectratio]{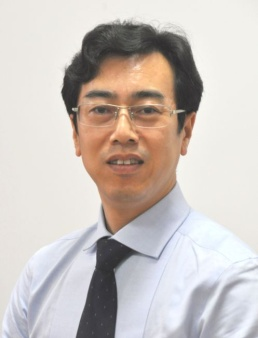}}]{Weidong Chen}
(M’04) received his B.S. and M.S. degrees in Control Engineering in 1990 and 1993, and Ph.D. degree in Mechatronics in 1996, respectively, all from the Harbin Institute of Technology, Harbin, China. Since 1996, he has been at the Shanghai Jiao Tong University where he is currently Professor of the Department of Automation, and Deputy Dean of the Institute of Medical Robotics. He is the founder of the Autonomous Robot Laboratory. From 2013 to 2019, he served as Chair of the Department of Automation. Dr. Chen’s current research interests include perception and control of robotic systems, multi-robot systems and medical robotics.
\end{IEEEbiography}

%\end{IEEEbiography}


%\newpage



%\vfill


\end{document}